\documentclass[sigconf,9pt,nonacm]{acmart}

\usepackage[english]{babel}
\usepackage{blindtext}
\usepackage{caption}
\usepackage{subcaption}
\usepackage{color}
\usepackage{tabularx}
\usepackage{multirow}
\usepackage{multicol}
\usepackage{graphicx}
\usepackage{tabularray}
\usepackage{wrapfig}

\usepackage[inline]{enumitem}
\setlist[itemize]{leftmargin=*, topsep=2pt, itemsep=1pt}

\usepackage{float}
\usepackage{xcolor}
\usepackage{todonotes}
\usepackage{array}
\usepackage{makecell}
\usepackage{booktabs}
\usepackage{geometry}
\usepackage{microtype}
\usepackage{xspace}

\newcommand{\eg}{e.g.\@\xspace}

\renewcommand\footnotetextcopyrightpermission[1]{}
\setcopyright{none}

\acmDOI{}
\acmISBN{}

\newcommand{\framework}{Chorus}

\colorlet{blue}{black}

\begin{document}

\title{\framework: Harmonizing Context and Sensing Signals for Data-Free Post-Deployment Model Customization in IoT}


\author{Liyu Zhang}
\affiliation{%
  \institution{Hong Kong University of Science and Technology}
  \country{Hong Kong SAR, China}
}
\email{lzhangcx@connect.ust.hk}

\author{Yejia Liu}
\affiliation{%
  \institution{Hong Kong University of Science and Technology}
  \country{Hong Kong SAR, China}
}
\email{yliutb@connect.ust.hk}

\author{Kwun Ho Liu}
\affiliation{%
  \institution{Hong Kong University of Science and Technology}
  \country{Hong Kong SAR, China}
}
\email{khliuae@connect.ust.hk}

\author{Runxi Huang}
\affiliation{%
  \institution{Hong Kong University of Science and Technology}
  \country{Hong Kong SAR, China}
}
\email{rhuangbj@connect.ust.hk}

\author{Xiaomin Ouyang}
\authornote{Corresponding author.}
\affiliation{%
  \institution{Hong Kong University of Science and Technology}
  \country{Hong Kong SAR, China}
}
\email{xmouyang@cse.ust.hk}

\renewcommand{\shortauthors}{Zhang et al.}


\begin{abstract}

A key bottleneck toward scalable IoT sensing is how to efficiently adapt trained AI models to new deployment conditions.
In real-world IoT systems, sensor data is collected under diverse and changing contexts, such as varying sensor placements or ambient environments, which can significantly alter signal patterns and degrade downstream performance.
Traditional domain adaptation and generalization methods often ignore such contextual information or model it in overly simplistic ways, limiting their ability to handle unseen context shifts under resource-constrained IoT deployment settings.

In this paper, we propose \framework{}, a \emph{context-bridged}, \emph{data-free post-deployment model customization} approach that efficiently learns context representations of new deployment environments to enable customization of sensing models, without requiring target-domain sensor data or post-deployment model retraining.

The key idea is to learn effective context representations that capture how contextual variations reshape sensor data patterns, and then exploit these representations as structured priors for robust inference-time model customization under unseen context shifts.

Specifically, \framework{} regularizes the context embedding space to yield compact, transferable representations, and further aligns it with the sensor latent space using unlabeled sensor--context pairs, explicitly bridging context generalization with sensing-data generalization.

Building on the regularized and aligned representations, \framework{} trains a lightweight gated prediction head with limited labeled data to effectively integrate context priors at inference, and introduces an adaptive caching mechanism that reuses previously cached context representations when no context shift is detected, thereby reducing on-device inference overhead.

Extensive experiments across IMU sensing, speech enhancement, and WiFi sensing tasks under diverse environmental context shifts show that \framework{} outperforms state-of-the-art baselines by up to 20.2\% in unseen contexts, achieves inference latency comparable to sensor-only deployment, and maintains stable performance under continuous real-world context transitions and varied phrases of context descriptions.

A video demonstration of \framework{}'s performance in the real world is available at
\url{https://youtu.be/yANTZsk0TVU}.

\end{abstract}

\maketitle


\section{Introduction}

 
 
 
 
 
 


\begin{figure}[t]
    \centering
    \setlength{\abovecaptionskip}{0.cm}
    \setlength{\belowcaptionskip}{0.cm}
    \includegraphics[width=\columnwidth]{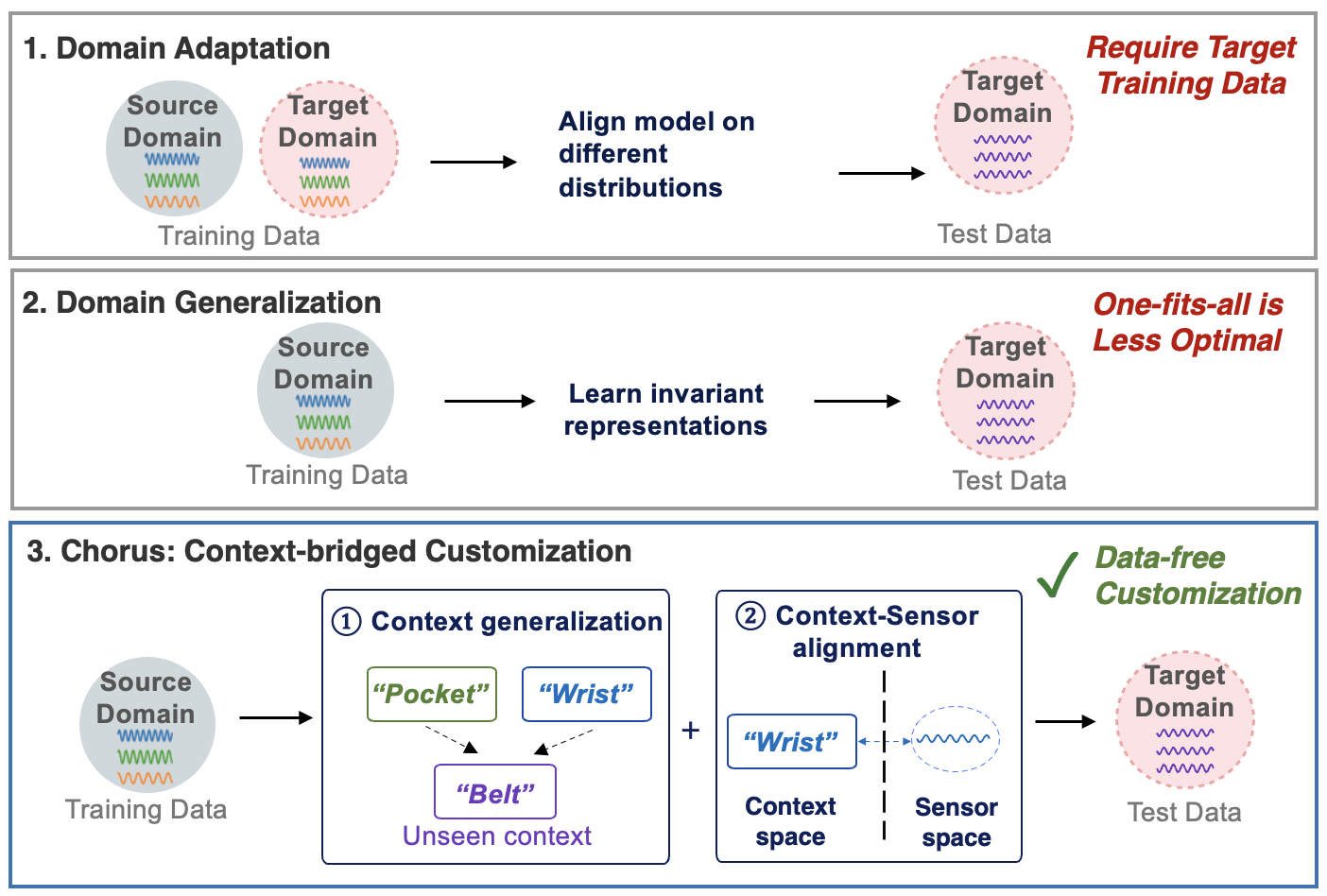}
    \caption{\framework{} learns effective context representations for new deployment settings and explicitly bridges context generalization with sensing generalization, enabling sensing-model customization without requiring target-domain sensor data or post-deployment model retraining.}
    \label{fig:intro_overview}
\end{figure}

{A key bottleneck toward scalable Internet of Things (IoT) sensing is how to efficiently adapt trained sensing models to new deployment conditions.}
Many IoT applications rely on heterogeneous sensors such as IMUs, microphones, and wireless radios to perceive the physical world for downstream analysis or control tasks~\cite{Satyanarayanan2017,hoseini2013smartphonecontext,miluzzo2008cenceme,vaizman2017context}. However, these sensors often operate under rapidly changing \emph{contexts}, where operational and environmental conditions induce substantial distribution shifts~\cite{korel2010survey,lane2015deepear,vaizman2017context}. 
For example, in wearable sensing, sensor placement encodes critical contextual information: an accelerometer trace differs markedly when a smartwatch is worn on the wrist versus clipped to the arm \cite{Sztyler2017PositionAware}. 
Similarly, in wireless sensing, Wi-Fi Channel State Information (CSI) signals vary significantly across physical environments, such as line-of-sight versus through-wall settings~\cite{Kotaru2015SpotFi,ma2019csisurvey}.

In machine learning, such physical \emph{context shifts} induce statistical \emph{domain gaps}, where data distributions vary substantially across deployment contexts~\cite{DomainAdaptation}. 
As illustrated in Figure~\ref{fig:intro_overview}, existing domain adaptation and generalization paradigms leave a critical gap for IoT sensing systems, where post-deployment data and model retraining are often unavailable or severely constrained. 
Domain adaptation improves performance by aligning source and target distributions, but it relies on access to target-domain training data. Domain generalization aims to learn invariant representations without target-domain samples; however, the one-size-fits-all representations may suppress context-specific sensor variations and therefore fail to achieve optimal performance across diverse deployment settings.
In real-world deployments, target-domain sensor data are often unavailable prior to deployment, whereas lightweight context information—such as sensor placement, environment type, or device condition—is typically accessible through system metadata, user configuration, or lightweight context detectors~\cite{wang2014studentlife,lee2023androidsurvey}. In contrast to prior approaches, \framework{} leverages only deployment-time context descriptions, without requiring any target-domain sensor samples, to efficiently customize sensing models by explicitly bridging context generalization and sensing-data generalization.


Although several recent approaches have explored efficient model customization after deployment, they still face important limitations for IoT sensing applications. Few-shot fine-tuning and test-time adaptation methods still rely on a few target-domain samples or deployment-time model updates, introducing non-trivial computational and memory overhead on resource-constrained mobile and IoT devices~\cite{gong2019metasense,pmlr-v70-finn17a,wang2021tent,liang2025tta_survey,tang2024adashadow,nazar2025}. This limitation becomes particularly problematic when the underlying sensing models continue to grow in size and deployment contexts change rapidly, requiring frequent model updates while making real-time performance hard to maintain.
Meanwhile, although several prior works incorporate contextual information into sensing systems, they often model context in overly simplistic ways, such as static side features, rule-based conditions, or expert selectors~\cite{miao2015context,zhou2025dgsense}. As a result, they fail to explicitly learn context representations that characterize how contextual variations reshape sensor data distributions, limiting their ability to generalize under unseen context shifts. As shown in Section~\ref{sec:motivate_study}, naive context integration can even degrade performance when raw context embeddings are not properly structured and aligned with shifts in the sensor latent space.

In this paper, we propose \framework{}, a new \emph{context-bridged}, \emph{data-free post-deployment model customization} framework that learns effective context representations of new deployment environments to enable sensing model customization without requiring target-domain sensor data or post-deployment retraining of the underlying sensing model. As illustrated in Figure~\ref{fig:intro_overview}, the key idea of \framework{} is to learn expressive context representations that capture how contextual variations influence sensor data patterns, and then exploit these representations as structured priors for robust and lightweight model customization under unseen context shifts.

Specifically, \framework{} is designed around two key objectives: effective context representation learning and efficient context integration. First, \framework{} regularizes the context embedding space to yield compact and transferable representations, and further aligns the context space with the sensor latent space through unlabeled sensor--context pairs. This design explicitly bridges context generalization and sensing-data generalization, enabling effective customization to unseen target contexts by leveraging both the transferability of context representations and their alignment with sensor distributions.
Building upon the regularized and aligned representations, \framework{} trains a lightweight gated prediction head using limited labeled data to enable effective context integration during inference. To further reduce inference overhead, \framework{} introduces an adaptive caching mechanism that reuses previously cached context representations when no context shift is detected, thereby minimizing additional inference latency and computation cost.

We evaluate \framework{} against 9 state-of-the-art baselines across IMU sensing, speech enhancement, and WiFi sensing tasks under 15 diverse context conditions. The results show that \framework{} consistently improves performance under unseen context shifts and with various paraphrased context descriptions, achieving up to 20.2\% performance gains over the strongest baseline.
Evaluations on multiple mobile devices further demonstrate that, with the adaptive caching design of the context encoder, the inference latency of \framework{} remains comparable to sensor-only deployment. We also evaluate \framework{} under continuous real-world context transitions and show that it maintains stable and robust performance across different context-shift phases.

In summary, we make the following key contributions:
\begin{itemize}
    \item We conduct an in-depth motivation study on context shifts in sensing applications, showing that model performance degrades substantially as context shift increases, while naive context integration strategies fail to generalize reliably across diverse unseen deployment contexts.
    \item We propose \framework{}, a new data-free post-deployment model customization framework for IoT sensing under unseen deployment contexts. \framework{} learns transferable and sensing-aligned context representations, and efficiently integrates them for inference-time model customization through a lightweight gated prediction head and an adaptive context caching mechanism.
    \item We evaluate \framework{} across IMU sensing, speech enhancement, and WiFi sensing tasks under 15 context conditions, showing consistent performance gains under unseen context shifts, low inference overhead, and stable performance under continuous real-world context transitions.
\end{itemize}

\section{Related Work}

\textbf{Domain Adaptation}. 
\textcolor{blue}{Relevant adaptation methods for sensing include unsupervised domain adaptation, such as adversarial adaptation methods represented by DANN~\cite{DANN} and sensing-specific systems such as M3BAT~\cite{meegahapola2024m3bat}, as well as few-shot adaptation (e.g., TS2ACT~\cite{TS2ACT}) and test-time adaptation (e.g., OFTTA~\cite{OFTTA}).}
\textcolor{blue}{Despite improving robustness, these methods remain target-dependent: they require target-domain samples, target labels, or deployment-time updates to adapt to new conditions.
This assumption is poorly matched to mobile and IoT sensing, where contexts can change rapidly and target-domain sensing data may be unavailable before use.
Test-time adaptation can further introduce latency and energy overhead because it often updates model parameters or statistics using deployment streams.
By contrast, \framework{} targets \emph{data-free post-deployment model customization} for unseen deployment contexts, without requiring target-domain sensor data or post-deployment retraining of the underlying sensing model.}

\textbf{Domain Generalization}. 
Domain generalization removes the need for target data by learning representations that transfer across unseen domains~\cite{DomainGeneralization}. 
\textcolor{blue}{Representative approaches often reduce domain-specific variation through invariance-inducing objectives~\cite{ContrasGAN}, and recent sensing systems such as DGSense~\cite{zhou2025dgsense} follow a similar invariance-driven principle. 
Source-only time-series representation learning methods, such as TS-TST~\cite{tst}, can also improve temporal feature extraction, but they do not explicitly model how deployment context reshapes sensor patterns.}
However, invariance-driven or generic source-only representations can inadvertently suppress context-specific cues that remain useful under strong shifts.
\textcolor{blue}{By contrast, \framework{} explicitly bridges context generalization with sensing-data generalization. It learns compact, transferable, and sensing-aligned context representations from unlabeled sensor--context pairs, and uses them as structured priors for lightweight inference-time customization under unseen context shifts.}

\textbf{Context-Aware Sensing}. 
\textcolor{blue}{Prior work in context-aware sensing has used contextual side information to improve robustness~\cite{ge2023dhc_hgl,vaizman2017context}, and CACTUS is a representative recent system in this direction~\cite{CACTUS}. 
However, these methods often incorporate context through static side features, fixed fusion, expert routing, or context-specific subnetworks. 
For example, CACTUS follows a specialist architecture in which model structure grows with the number of context-specific experts, increasing parameter size and edge cost as contexts expand. 
More importantly, such designs do not explicitly learn context representations that capture how contextual variations reshape sensor data patterns. 
By contrast, \framework{} treats context as a learnable representation space rather than a fixed side input. It regularizes and aligns context representations with the sensor latent space, then integrates the resulting context priors through a lightweight gated prediction head and adaptive context cache, enabling data-free post-deployment customization without growing context-specific expert branches.}

\section{A Motivation Study}
\label{sec:motivate_study}

\subsection{Performance under Context Shifts}
\label{sec:motivation1}
\textcolor{blue}{We first examine how deployment context shifts affect sensing performance and why context descriptions must be transformed into effective representations before they can serve as structured priors for sensing-model customization.}
To systematically analyze this challenge, we evaluate model performance using IMU data from 10 subjects in the public Shoaib dataset~\cite{shoaib} for human activity recognition (HAR). 
The task is to classify six human activities across different on-body sensor placements (e.g., \textit{Wrist}, \textit{Left pocket}, \textit{Right pocket}, \textit{Upper Arm}, and \textit{Belt}).

\textcolor{blue}{We use a latent MMD metric~\cite{gretton2012kernel} as a sensor-space proxy for the distribution gap induced by context shifts.}
\textcolor{blue}{The metric is computed on sensor representations extracted by a source-only vanilla autoencoder trained only with reconstruction loss.}
\textcolor{blue}{A higher latent MMD score indicates a larger sensor-space gap caused by the context shift.}
Using ``Left pocket'' and ``Right pocket'' as the source training contexts, we categorize three levels of shift for the remaining placements: \emph{low} (``Wrist'', latent MMD $= 0.587$), \emph{mid} (``Upper Arm'', latent MMD $= 0.873$), and \emph{high} (``Belt'', latent MMD $= 1.571$).

\begin{figure}[t]
    \centering
    \setlength{\abovecaptionskip}{0.cm}
    \setlength{\belowcaptionskip}{-0.cm}
    \begin{subfigure}[t]{0.53\linewidth}
        \centering
        \includegraphics[width=\linewidth]{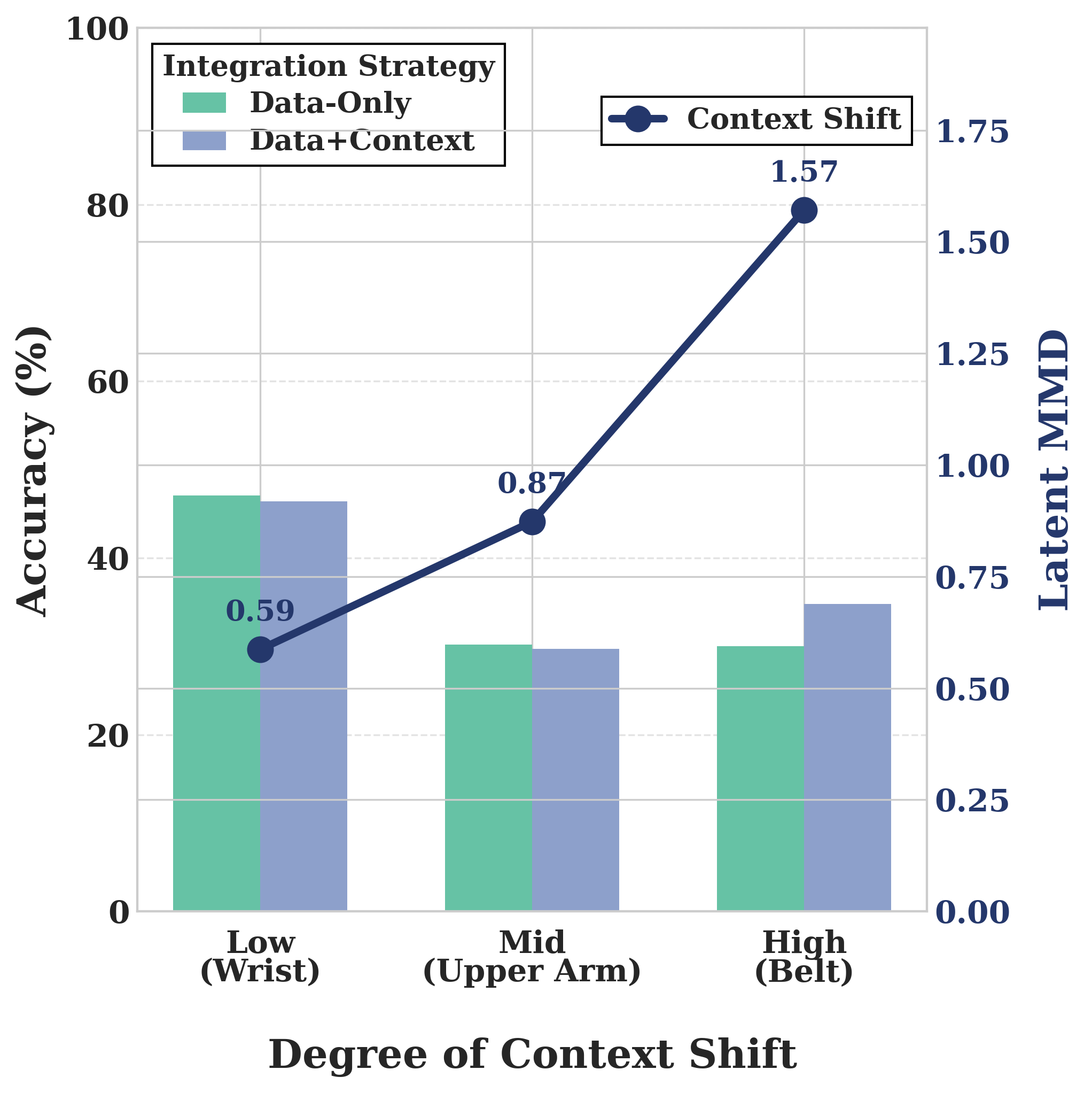}
        \caption{Performance across context shifts.}
        \label{fig:motivation_shift}
    \end{subfigure}%
    \hfill
    \begin{subfigure}[t]{0.45\linewidth}
        \centering
        \includegraphics[width=\linewidth]{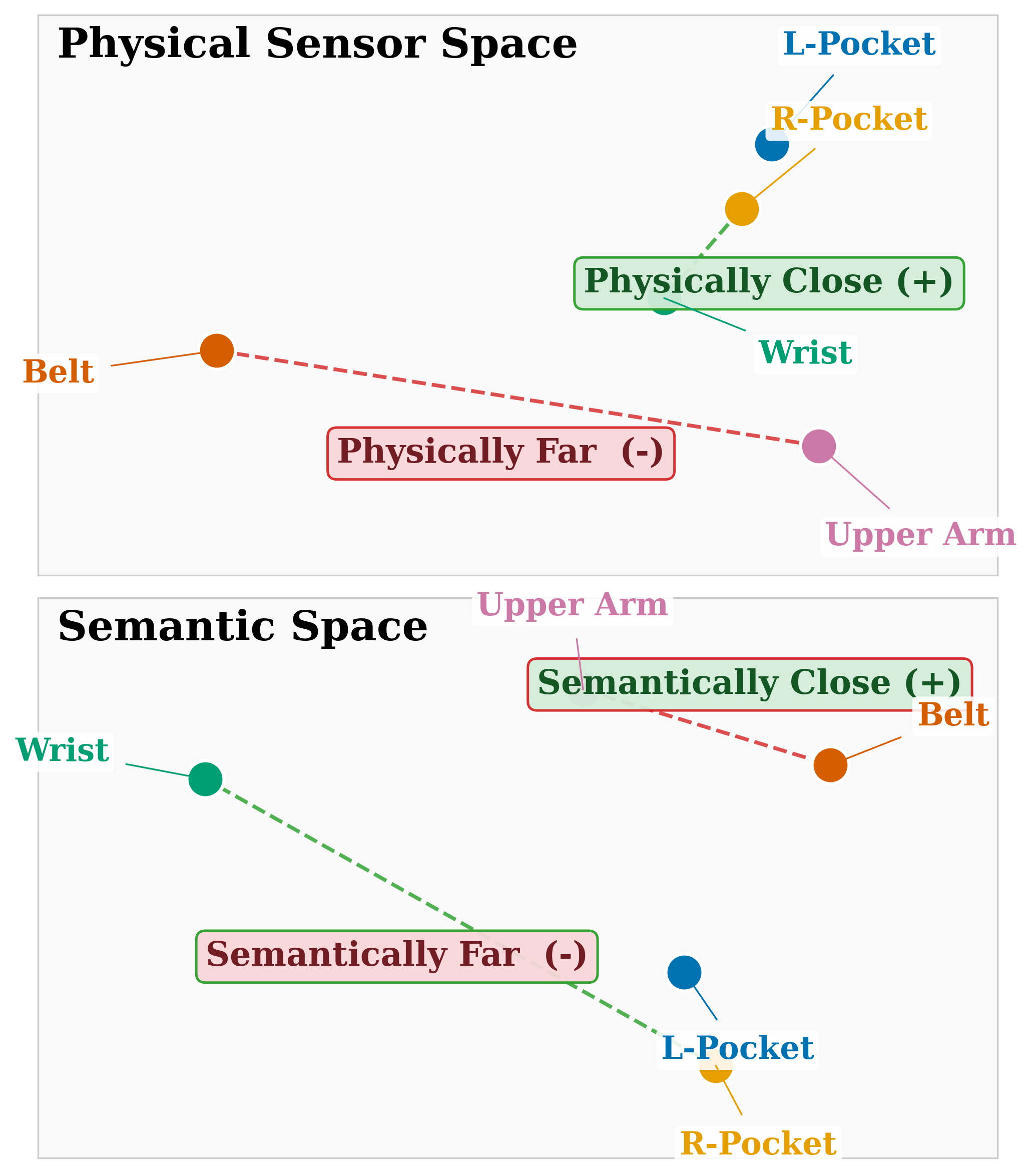}
        \caption{\textcolor{blue}{Mismatch between sensor and context spaces.}}
        \label{fig:motivation_topology}
    \end{subfigure}
    \caption{\textcolor{blue}{Motivation study. Context shifts degrade sensing performance, while naive context injection is unstable because raw semantic context embeddings are not compact or sensing-aligned representations of physical context variations.}}     
    \label{fig:motivation_figures}
\end{figure}

As a reference, we first consider a \emph{Data-Only} approach that predicts activities solely from sensor inputs without using any explicit context information. 
As shown by the Data-Only bars and the latent MMD curve in Figure~\ref{fig:motivation_figures}(a), the Data-Only model drops sharply as the sensor-space gap induced by context shifts grows from Wrist to Upper Arm and remains low under the high-shift Belt setting.
\textcolor{blue}{To address this degradation, a straightforward idea is to integrate deployment context descriptions into the model.}
We compare the \emph{Data-Only} approach against a \emph{Data+Context} strategy, which integrates pre-trained language embeddings of the placement descriptions (via MiniLM-v2~\cite{wang2020minilm}) using simple addition.

However, naively injecting context does not uniformly improve accuracy. In the low-shift \textit{Wrist} scenario, adding context slightly reduces accuracy from 47.1\% to 46.4\%. 
A similar pattern appears in the mid-shift \textit{Upper Arm} scenario, where accuracy drops from 30.2\% to 29.7\%.
Only in the high-shift \textit{Belt} context does naive context fusion become beneficial, improving accuracy from 30.0\% to 34.8\%.  
\textcolor{blue}{This pattern shows that deployment context can provide useful priors, but overly simple context incorporation is unstable because raw context embeddings are not necessarily compact, transferable, or aligned with sensor-space variations induced by physical contexts.}

\subsection{Understanding Context Integration}
\label{sec:motivation3}
To understand why naive context injection behaves so inconsistently, we compare the pairwise relationships among the five placement contexts in two independent spaces. 
In the latent sensor space, we measure pairwise distances using latent MMD between context-specific IMU representations extracted by a source-only vanilla autoencoder. In the semantic context space, we encode the same placement labels using MiniLM-v2~\cite{wang2020minilm}, compute pairwise cosine distances, and project both distance matrices with multidimensional scaling (MDS)~\cite{cox2001mds}.

Figure~\ref{fig:motivation_figures}(b) reveals a clear topological mismatch between these two spaces. Contexts that are physically similar can be pushed far apart in the semantic space, while physically distinct contexts can be grouped together by linguistic similarity. For example, \textit{Wrist} remains relatively close to the pocket placements in the latent sensor space, consistent with their shared limb-motion dynamics, yet it is placed much farther away in the semantic space. In contrast, \textit{Upper Arm} is noticeably farther from the pocket placements and \textit{Belt} is the most distant in the latent sensor space, while the semantic embedding still tends to group these coarse-grained body-worn locations together.
\textcolor{blue}{As a result, the raw semantic prior can distort the physical relationships among contexts that the sensing model should rely on. 
When context is injected through naive addition, the model must fuse misaligned representations, which leads directly to the unstable behavior observed in Figure~\ref{fig:motivation_figures}(a). 
This suggests that context descriptions should first be regularized and aligned with sensor-space variations before being used as structured priors for customization.}

\subsection{Summary}
\textcolor{blue}{Our motivation study highlights two key insights that drive the design of \framework{}:}
\begin{itemize}
    \item \textcolor{blue}{The performance of sensing models degrades under larger sensor-space gaps induced by context shifts, dropping sharply from low- to mid-shift settings and remaining low under high-shift settings. Although context can provide useful priors, naive integration strategies such as simple additive fusion remain unreliable under unseen contexts.}
    \item \textcolor{blue}{Raw text embeddings of context descriptions exhibit a severe topological mismatch with sensor embeddings: physically similar contexts may appear far apart in semantic space, while physically distinct contexts may be spuriously clustered together.}
\end{itemize}

\textcolor{blue}{Therefore, robust data-free post-deployment model customization requires effective context representation learning that bridges context generalization and sensing-data generalization.
This motivates \framework{} to map deployment-time context descriptions into a compact, transferable, and sensing-aligned space, so that the learned context representations can serve as structured priors under unseen context shifts.
Once such representations are learned, \framework{} further needs efficient context integration to use these priors with low inference overhead.}

\section{System Overview}
\label{sec:system_overview}

\subsection{Applications and Challenges}
\framework{} targets IoT sensing applications in which sensor data is collected under diverse and dynamically changing contextual conditions, requiring the system to optimize sensing model performance after deployment without retraining, due to limited on-device computing resources. Here, \emph{context}~\cite{dey2001context} refers to environmental and operational factors, such as device placement, ambient noise, or user mobility, that systematically affect sensor patterns.

In human activity recognition with wearables, on-body sensor placement (\eg, wrist vs. ankle) is a critical context that substantially changes IMU signal patterns. In speech enhancement, environmental context, such as a quiet office versus a busy street, introduces different ambient noise patterns in raw audio. In wireless sensing, user mobility and physical environment significantly affect Wi-Fi Channel State Information (CSI), leading to context-dependent multipath fading and Doppler effects.

\textcolor{blue}{Across these applications, contextual variation raises two design challenges. First, effective context representation learning is needed: context descriptions should be transformed into compact, transferable, and sensing-aligned representations that capture how contextual variations reshape sensor data patterns, so that context generalization can support unseen deployment conditions. Second, efficient context integration is needed: many IoT sensing tasks are latency-sensitive, requiring context priors to be integrated with streaming sensor representations and reused without costly deployment-time updates.}

\begin{figure}
  \centering
  \setlength{\abovecaptionskip}{0.cm}
  \setlength{\belowcaptionskip}{0.cm}

  \begin{subfigure}[b]{0.15\textwidth}
    \centering
    \includegraphics[width=\linewidth]{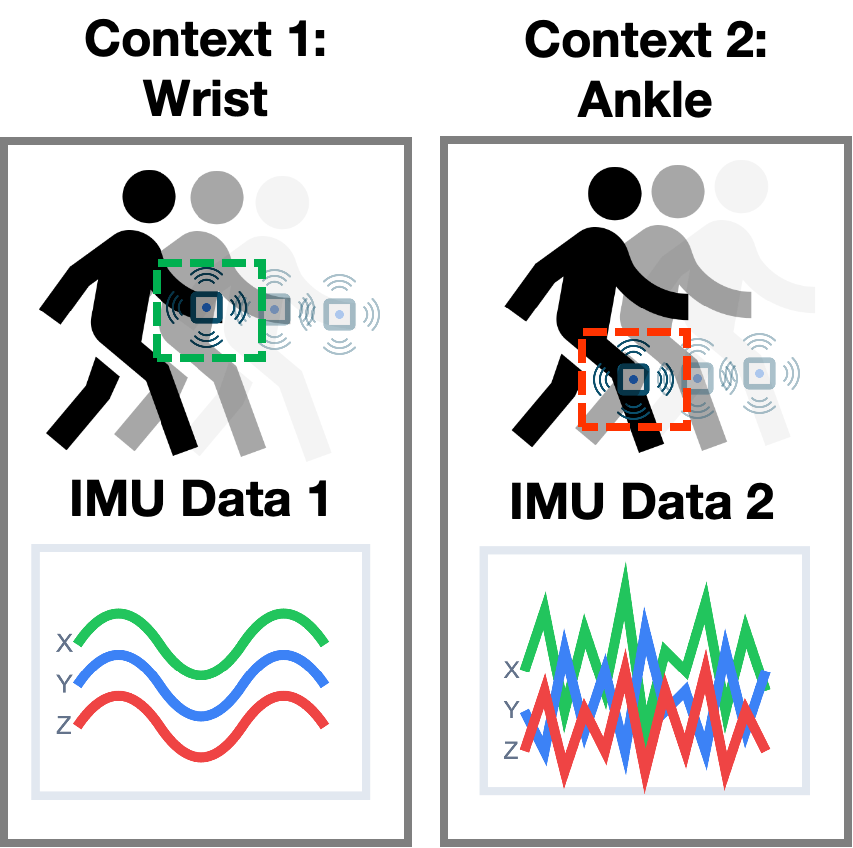}
    \vspace{2pt}
    \scriptsize
    (a) IMU-based HAR\\
    (Context: placement)
  \end{subfigure}
  \hfill
  \begin{subfigure}[b]{0.15\textwidth}
    \centering
    \includegraphics[width=\linewidth]{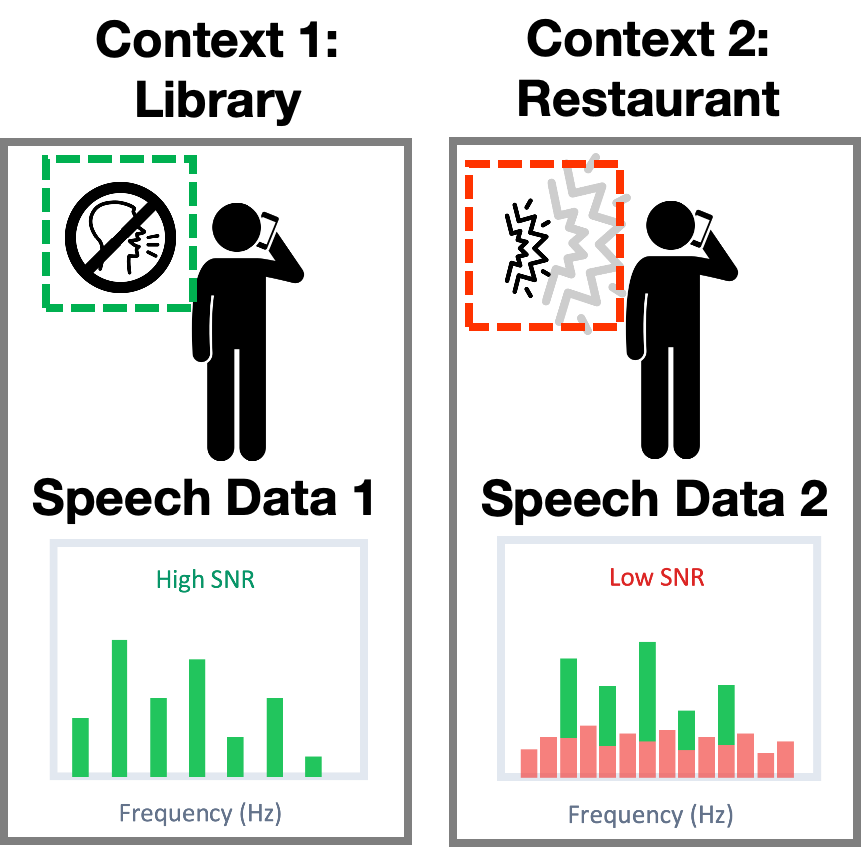}
    \vspace{2pt}
    \scriptsize
    (b) Speech enhancement\\
    (Context: environment)
  \end{subfigure}
  \hfill
  \begin{subfigure}[b]{0.15\textwidth}
    \centering
    \includegraphics[width=\linewidth]{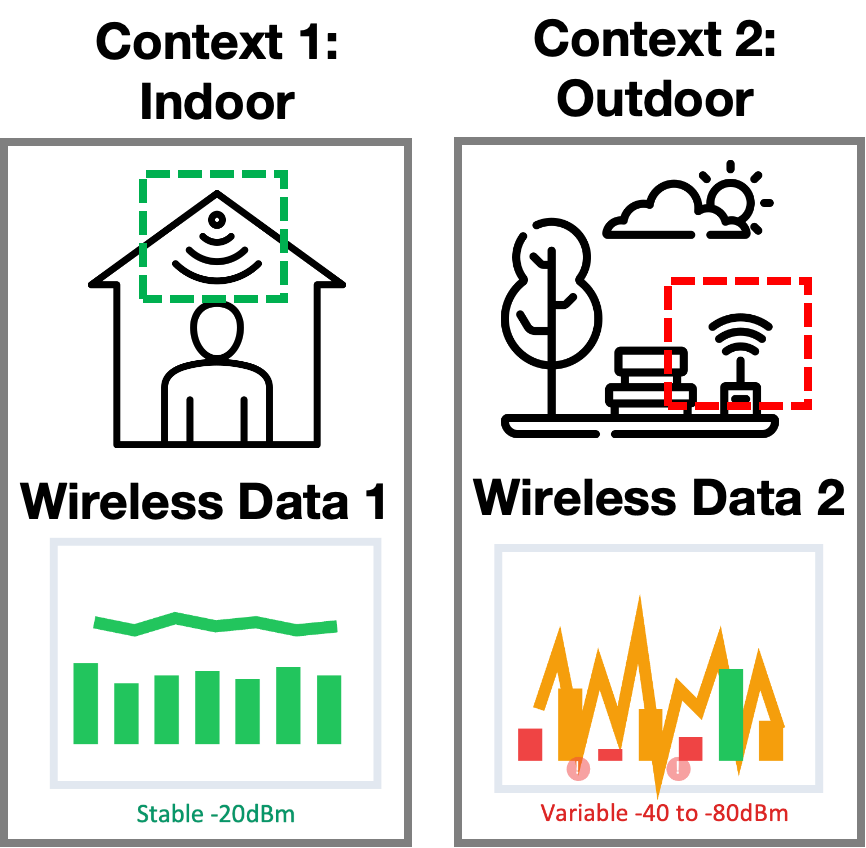}
    \vspace{2pt}
    \scriptsize
    (c) WiFi for fall detection\\
    (Context: environment)
  \end{subfigure}

  \vspace{4pt}
  \caption{Context shifts in different sensing applications.}
  \label{fig:app_scenarios}
\end{figure}

\subsection{Problem Formulation}
\textcolor{blue}{The goal of \framework{} is to enable \emph{context-bridged}, \emph{data-free post-deployment model customization} under target contexts unseen during training. 
During training, \framework{} uses unlabeled source sensor--context pairs collected from seen contexts for context representation learning, and a small amount of labeled source sensor--context data for training the lightweight gating module and task head. 
Each context is represented by a lightweight description, such as a placement tag, room type, or device configuration, rather than by target-domain sensing samples.}

\textcolor{blue}{At deployment time, the target context may be unseen during training, but its context description can be obtained from metadata, user setup, or a lightweight context detector. 
The data-free post-deployment setting means that \framework{} does not use target-domain sensing samples, target labels, or post-deployment model updates to customize the model. 
Instead, it uses the target context description as a bridge: the description is mapped into the learned context representation space and then integrated with streaming sensor representations for prediction. 
}

\subsection{System Architecture}
\begin{figure*}[t]
    \centering
    \setlength{\abovecaptionskip}{0.cm}
    \setlength{\belowcaptionskip}{0.cm}
    \includegraphics[width=\textwidth]{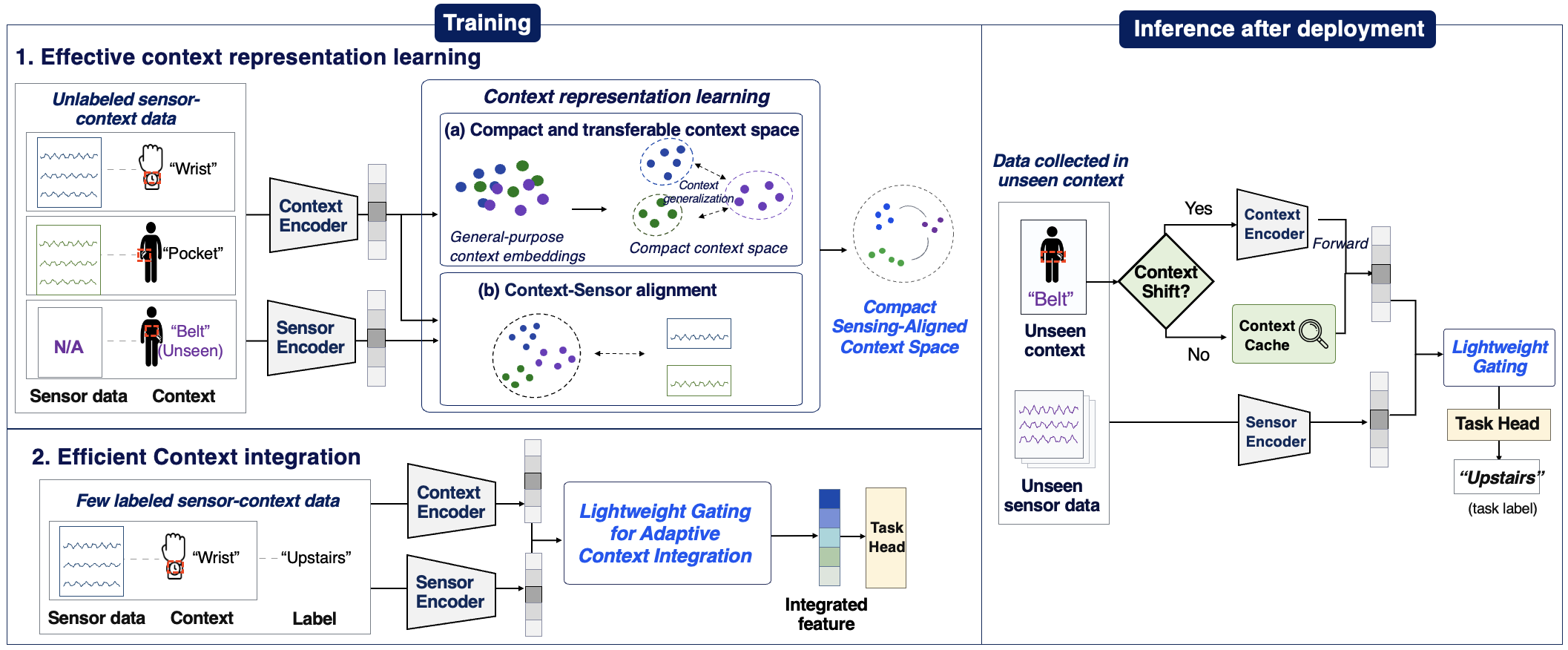}
    \caption[System overview]{%
    \textcolor{blue}{System overview. \framework{} performs data-free post-deployment customization through effective context representation learning and efficient context integration. During training, it learns a compact sensing-aligned context space from unlabeled sensor--context pairs and trains a lightweight gating module with limited labeled source data. During inference, it projects an unseen context description into the learned space, integrates the context prior with streaming sensor representations, and reuses cached context representations until the deployment context changes.}
    }
    \label{fig:system_overview}
\end{figure*}
\textcolor{blue}{The design of \framework{} is motivated by the key insight from Section~\ref{sec:motivate_study}: raw context descriptions do not naturally capture how physical contexts reshape sensor data patterns. 
As shown in Figure~\ref{fig:system_overview}, \framework{} therefore treats context as a bridge between context generalization and sensing-data generalization, enabling data-free post-deployment customization without target-domain sensor data or post-deployment retraining. 
This key idea leads to two design goals: effective context representation learning and efficient context integration for model customization.}

\textcolor{blue}{The first design component is effective context representation learning. 
Given unlabeled source sensor--context pairs from seen deployment contexts, \framework{} uses a context encoder and a sensor encoder to learn a compact, transferable, and sensing-aligned context space. 
The context branch first regularizes general-purpose language context embeddings into a compact context space, so that semantically related context descriptions can be projected into a stable representation region and unseen context descriptions can be generalized into this space. 
\framework{} then aligns the compact context space with the sensor representation space using unlabeled but paired sensor--context data, so that the learned context representations reflect how contextual variations reshape sensor patterns. 
This design turns deployment-time context descriptions into structured priors that can be used even when the corresponding target-domain sensor distribution has not been observed during training.}

\textcolor{blue}{The second design component is efficient context integration. 
After learning the sensing-aligned context space, \framework{} freezes the learned context and sensor encoders and trains a lightweight gating module together with a task head using few labeled sensor--context samples from source domains. 
The gating module learns how to combine the sensor representation with the learned context prior at the sample level, instead of relying on fixed addition, concatenation, or context-specific expert branches. 
This keeps the context integration module lightweight while allowing the model to adjust how much it relies on sensor evidence or context prior for each streaming sensor segment.}

\textcolor{blue}{During post-deployment inference, \framework{} receives streaming sensor data and a deployment-time context description for a possibly unseen context. 
When a context shift is detected, the context description is forwarded through the context encoder and projected into the learned compact and transferable sensing-aligned context space. 
When no context shift is detected, \framework{} performs a cache lookup and reuses the previously cached context representation to bypass  context encoding. 
The sensor encoder processes each streaming sensor segment, and the lightweight gating module integrates the sensor representation with the cached or newly computed context prior before the task head produces the final prediction. 
In this way, \framework{} performs lightweight inference-time sensing model customization using only the target context description, while avoiding target-domain sensor samples, post-deployment model updates, and repeated context encoding under stable deployment contexts.}


\section{Design of \framework{}}
\label{sec:design}

\textcolor{blue}{\framework{} is designed around two components: effective context representation learning and efficient context integration. 
First, \framework{} regularizes raw context embeddings into compact and transferable context representations that can generalize to unseen contexts, and then aligns these representations with the sensor space using unlabeled sensor--context pairs. 
Second, \framework{} integrates the learned sensing-aligned context representations as structured priors for post-deployment customization through a lightweight gating module, while reusing cached context representations when the deployment context remains stable to reduce on-device overhead.}

\subsection{\textcolor{blue}{Effective and Transferable Context Representation Learning}}
\label{sec:pretrain}

\textcolor{blue}{The first component of \framework{} is effective context representation learning. 
General language embeddings provide useful semantic cues, but they are not optimized to capture how physical contexts reshape sensor data patterns. 
\framework{} therefore first regularizes raw context embeddings into a compact and transferable context space using KL compactness regularization and context-identity contrastive separation. 
It then aligns this compact context space with the sensor space using unlabeled sensor--context pairs. This ordering is important: compact context representations provide the basis for context generalization, while context-sensing alignment makes this generalization useful for sensing-data generalization.}

Suppose $(x_s, c_s)$ denotes an unlabeled sensor--context pair from the source domains, where $x_s$ is a sensor segment and $c_s$ is its associated pre-computed context embedding derived from a short language description. 
For example, in HAR, this context may be an on-body placement such as ``Left pocket'' or ``Wrist'', while in speech enhancement it may be an acoustic environment such as ``Kitchen'' or ``Bus''. 
We map these two inputs into latent representations as
\[
z_x = f_{sensor}(x_s), \qquad
z_c = f_{context}(c_s).
\]
Here $f_{sensor}$ is a modality-specific encoder, such as a 1D CNN for IMU signals or a Transformer/CNN encoder for speech and WiFi features, while $f_{context}$ is a lightweight context adapter built on top of pre-computed MiniLM-L12-v2 sentence embeddings~\cite{wang2020minilm}.

\subsubsection{Compact and transferable context representation space.}
As shown in Figure~\ref{fig:motivation_figures}(b), raw semantic context embeddings do not necessarily reflect sensor-space variation across physical contexts. 
Directly reusing these pre-trained language embeddings therefore limits generalization to unseen contexts. 
\textcolor{blue}{To address this issue, \framework{} first regularizes the context latent space to encourage compact and transferable context representations.}
Specifically, \framework{} uses a KL term and a contrastive context-separation term to regularize the context latent space. 
The KL term regularizes the empirical distribution of context representations toward a compact prior:
\[
\mathcal{L}_{\text{KL}}
= D_{\text{KL}}\!\left(q(z_c)\,\|\,\mathcal{N}(0,I)\right),
\]
where $q(z_c)$ denotes the empirical distribution of context representations produced by $f_{\text{context}}$.
This term prevents raw language embeddings from spreading along broad semantic directions that are irrelevant to sensing, and encourages context representations to occupy a compact and stable latent region.
The separation term $\mathcal{L}_{\text{sep}}$ is implemented as an InfoNCE-style contrastive loss over context identities: it treats samples with identical context IDs as positives and samples from different context IDs in the mini-batch as negatives. 
Given an anchor context representation $z_i$, a positive representation $z_i^+$ from the same context identity, and negatives $\mathcal{N}(i)$ from different context identities, we define:
\[
\mathcal{L}_{\text{sep}}
= - \log
\frac{\exp(\mathrm{sim}(z_i,z_i^+)/\tau)}
{\exp(\mathrm{sim}(z_i,z_i^+)/\tau)
+ \sum_{z_j \in \mathcal{N}(i)}
\exp(\mathrm{sim}(z_i,z_j)/\tau)} ,
\]
where $\mathrm{sim}(\cdot,\cdot)$ denotes cosine similarity and $\tau$ is the temperature.
Its role is to keep different physical contexts distinguishable and prevent trivial collapse, where multiple contexts are mapped into an overly similar region.

\textcolor{blue}{This compact context space is the foundation for unseen-context generalization: it allows context descriptions to be represented in a stable and separable form before they are grounded in sensor-space variation.}

\subsubsection{\textcolor{blue}{Context-sensing alignment.}}
\textcolor{blue}{Compact context representations alone are not sufficient, because the context space also needs to reflect how sensor patterns vary under physical contexts. 
To make context generalization useful for sensing-data generalization, \framework{} aligns the compact context space with the sensor space using unlabeled sensor--context pairs.}

We combine bidirectional cross-modal reconstruction with within-modal self-reconstruction, so that the latent variables preserve modality fidelity while still enforcing cross-modal correspondence. 
Let $\hat{x}_{x \to x}$ and $\hat{c}_{c \to c}$ denote self-reconstructed sensor and context features, and let $\hat{x}_{c \to x}$ and $\hat{c}_{x \to c}$ denote cross-reconstructed features predicted from the opposite modality. 
We optimize:
\begin{align}
\mathcal{L}_{\text{recon}}
=&\ \alpha_{x \to x}\, d\!\left(\hat{x}_{x \to x}, x_s\right)
+ \alpha_{c \to c}\, d\!\left(\hat{c}_{c \to c}, c_s\right) \notag\\
&\ + \alpha_{c \to x}\, d\!\left(\hat{x}_{c \to x}, x_s\right)
+ \alpha_{x \to c}\, d\!\left(\hat{c}_{x \to c}, c_s\right).
\end{align}
Here $d(\cdot,\cdot)$ denotes a standard reconstruction distance between continuous targets. 
We use this generic form because both the sensor targets and the context targets are continuous tensors rather than discrete labels; in practice, it is instantiated with standard element-wise regression losses.
Such reconstruction objectives are widely used in multimodal generative models to align heterogeneous modalities in a shared latent space~\cite{ngiam2011multimodal, suzuki2016joint, shi2019variational, yi2021cross}. 
\textcolor{blue}{In our setting, this auxiliary objective helps preserve cross-modal correspondence between context and sensor representations. 
Together with context regularization and separation, it encourages the learned context space to reflect systematic sensor-pattern variations induced by physical contexts, such as how the same activity is expressed on the ``wrist'' versus the ``belt''.}

The full first-stage objective combines reconstruction, latent regularization, and contrastive context separation:
\[
\mathcal{L}_{\text{pre}}
= \mathcal{L}_{\text{recon}}
+ \beta\,\mathcal{L}_{\text{KL}}
+ \lambda_{\text{sep}}\,\mathcal{L}_{\text{sep}}.
\]
\textcolor{blue}{Together, these terms produce compact, transferable, and sensing-aligned context representations. 
In implementation, we may include a small number of stabilization terms for specific modalities, but the core representation-learning objective is governed by KL compactness regularization, context-identity contrastive separation, and alignment over unlabeled sensor--context pairs.}
\begin{figure}
    \centering
    \setlength{\abovecaptionskip}{0.cm}
    \setlength{\belowcaptionskip}{0.cm}
    \includegraphics[width=\columnwidth]{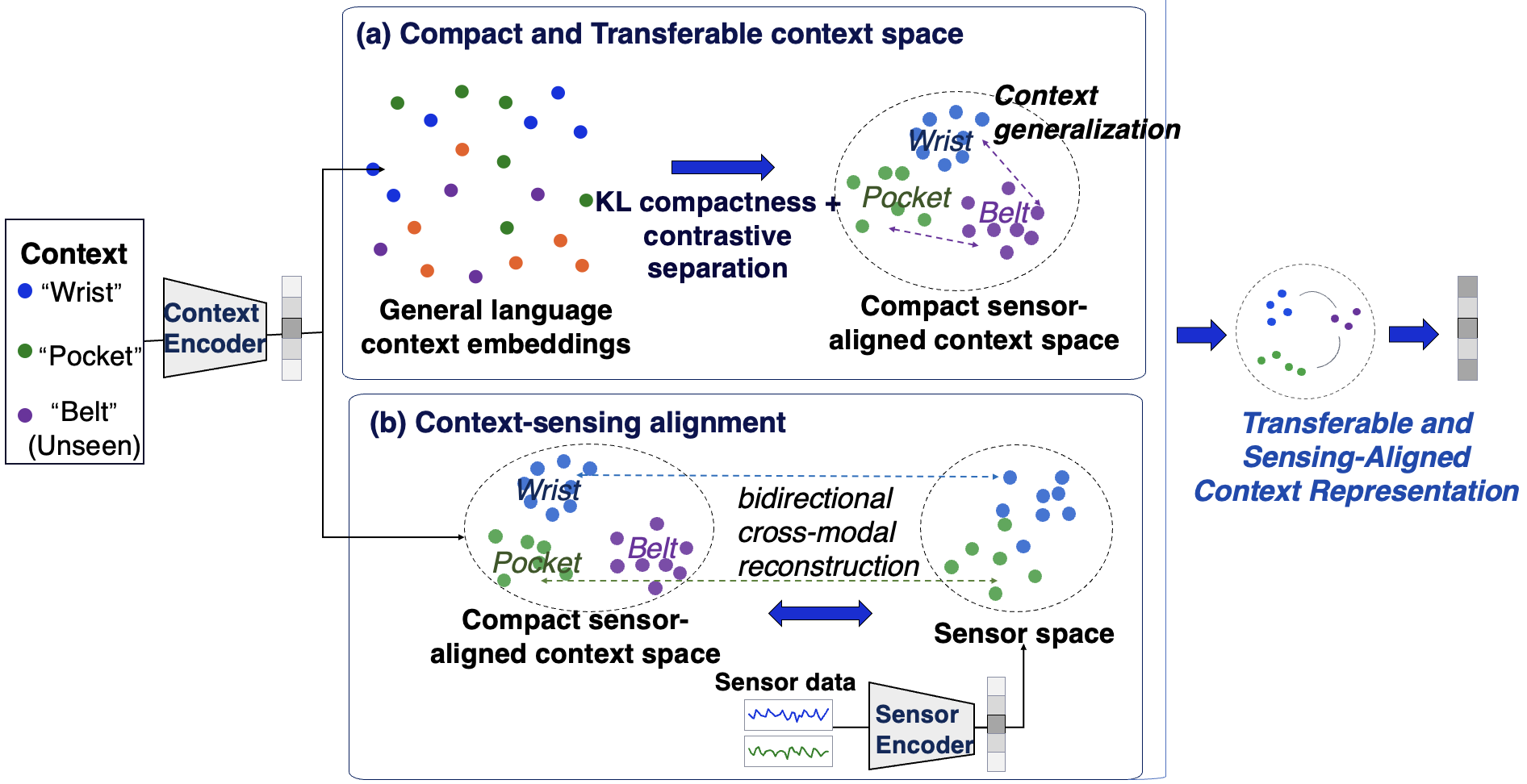}
    \caption{\textcolor{blue}{Effective context representation learning. \framework{} learns compact, transferable, and sensing-aligned context representations from raw context embeddings and unlabeled sensor--context pairs, enabling generalization to unseen contexts.}}
    \label{fig:cloud}
\end{figure}
\textcolor{blue}{As a result, the pre-trained $f_{\text{sensor}}$ captures both task structure and context-induced sensor variation, while the pre-trained $f_{\text{context}}$ produces transferable and sensing-aligned context representations. 
These learned context representations are then used as structured priors by the efficient context integration module.}
\textcolor{blue}{At deployment time, an unseen target context does not require any target-domain sensing samples to obtain a usable context prior. 
\framework{} encodes the target context description using the same context adapter and maps it into the learned compact context space. 
Because this space has been regularized to be compact and separable, and grounded in sensor-space variation during source training, the resulting target-context representation can be used as a structured prior even when the corresponding sensor distribution has not been observed during training. 
This is the mechanism by which context generalization is bridged with sensing-data generalization for post-deployment sensing-model customization under unseen deployment conditions.}


\subsection{\textcolor{blue}{Efficient Context Integration for Model Customization}}
\label{sec:adaptive_head}

\begin{figure}[t]
    \centering
    \setlength{\abovecaptionskip}{0.cm}
    \setlength{\belowcaptionskip}{0.cm}
    \includegraphics[width=\columnwidth]{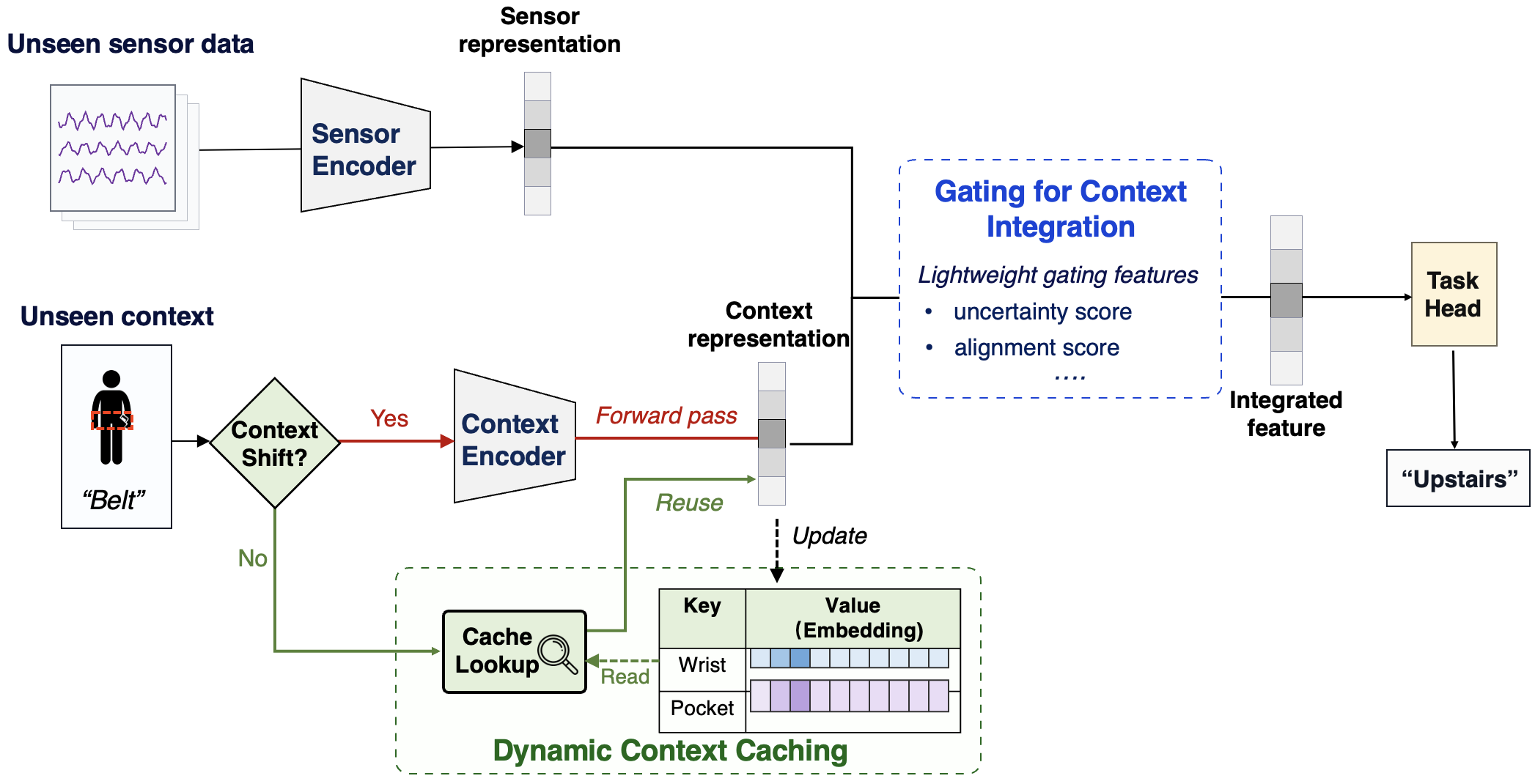}
    \caption{\textcolor{blue}{Efficient context integration. \framework{} trains a lightweight gating module to combine learned context priors with streaming sensor representations, and uses dynamic context caching to avoid repeated context encoding under stable deployment contexts.}}
    \label{fig:adaptive_head}
\end{figure}

\textcolor{blue}{The second component of \framework{} is efficient context integration for post-deployment customization. 
After effective context representation learning, \framework{} freezes the learned sensor and context encoders and trains a lightweight gating module and task head on limited labeled source sensor--context data. 
As shown in Figure~\ref{fig:adaptive_head}, this module contains two explicit blocks: a lightweight gating module that learns how to integrate the learned context prior with each streaming sensor representation, and dynamic context caching that reuses learned context representations across stable deployment contexts.
The gating module is the training-time integration design, while dynamic context caching is the post-deployment inference optimization.
Together, they allow \framework{} to use context priors effectively without target-domain sensing samples, post-deployment model updates, or repeated context encoding.}

\textcolor{blue}{This creates a two-stage training and deployment process. 
The representation stage uses unlabeled source sensor--context pairs to learn the compact sensing-aligned context space. 
The integration stage uses limited labeled source sensor--context data to train only the lightweight gating module and task head for the downstream task. 
During post-deployment inference, \framework{} does not update model parameters; it only refreshes cached context representations when the deployment context changes.}

\subsubsection{Lightweight gated integration.}
\textcolor{blue}{The lightweight gating module learns how to integrate the learned context prior with each streaming sensor representation. 
It is lightweight because it does not learn a new context representation, re-encode context descriptions, or update model parameters during deployment. 
Instead, it computes a compact set of lightweight gating features from the frozen sensor representation, the context prior, and simple statistics of the current sensor segment. 
These features allow the model to decide whether the current prediction should rely more on sensor evidence or on the learned context prior.}

Given a streaming sensor representation $h_{\text{sensor}}$ and a context representation $h_{\text{context}}$, the gated head builds a compact gating feature vector $r(x)$ from three types of lightweight cues. 
First, \emph{alignment} cues measure the compatibility between the current sensor representation and the context prior, capturing whether the learned context prior is consistent with the observed sensor segment. 
Second, \emph{dynamics} cues summarize simple statistics of the current input segment, such as short-window signal variation, which helps identify segments whose patterns are more sensitive to context shifts. 
Third, \emph{reliability} cues estimate the confidence or stability of the sensor branch for the current segment. 
In implementation, these cues are computed from lightweight feature interactions and statistics rather than from additional encoders. 
Because different sensing modalities exhibit context shifts in different ways, \framework{} can activate a modality-appropriate subset of these cues through a simple feature mask while keeping the same gated-head structure.

A small MLP $g_c$ maps the gating feature vector to two branch weights:
\[
\boldsymbol{\alpha}(x) =
[\alpha_{\text{sensor}}(x), \alpha_{\text{context}}(x)]
= \mathrm{softmax}(g_c(r(x))).
\]
The final representation is
\[
h_{\text{final}} =
\alpha_{\text{sensor}} h_{\text{sensor}}
+
\alpha_{\text{context}} h_{\text{context}}.
\]
A task-specific prediction head then produces the output $\hat{y}$.

\textcolor{blue}{This sample-level weighting is more flexible than fixed addition or concatenation. 
When the sensor pattern is reliable, the gated head can assign more weight to the sensor branch; when the sensor evidence becomes ambiguous under a context-induced shift, it can increase the influence of the learned context prior. 
Thus, the gated head provides effective context integration using only lightweight alignment, dynamics, and reliability signals.}

\textbf{Training objective.}
The inference module is trained on labeled source data with the task loss:
\[
\mathcal{L}_{\text{custom}} =
\mathcal{L}_{\text{task}}(y,\hat{y})
+
\lambda_{\text{balance}}\mathcal{L}_{\text{balance}}.
\]
The balance term discourages the gated head from always selecting the same branch, but it is lightweight and only used during training.

\subsubsection{Dynamic context caching.}
\label{sec:context_cache}

\textcolor{blue}{After the gating module is trained, dynamic context caching reduces the inference overhead of post-deployment customization. 
The key observation is that the context representation used by the gating module often remains stable across many consecutive sensor segments. 
Therefore, \framework{} computes the context representation only when a context shift is detected, updates the cache, and reuses the cached representation otherwise.}

To mitigate the computational overhead of repeatedly encoding context features during inference, we exploit the high temporal locality of context information in real-world streaming scenarios. 
In typical IoT deployments, contextual descriptors and their resulting embeddings often remain stable across extended sequences of input samples. 
For example, device placement usually does not change within a session, acoustic environments often persist for minutes, and wireless environments evolve much more slowly than packet-level traffic.

Let the streaming input at time step $t$ be $(x_t, u_t)$, where $x_t$ is the current sensor sample and $u_t$ is the associated context identifier, such as a placement label or environment tag. 
\framework{} maintains the cached context representation from the previous step and updates it only when the context identifier changes:
\[
z^{c}_{t} =
\begin{cases}
z^{c}_{t-1}, & \text{if } u_t = u_{t-1},\\
f_{\text{context}}(c(u_t)), & \text{if } u_t \neq u_{t-1}.
\end{cases}
\]
That is, if the current context matches the cached one, \framework{} directly reuses the previous latent context feature; otherwise, it recomputes the context representation and refreshes the cache. 
This changes context encoding from a per-sample operation to an event-driven update. 
For a stream with $T$ samples and only $M$ context changes, the number of context-encoding operations is reduced from $T$ to $M+1$.

\textcolor{blue}{Let $C_x$, $C_c$, and $C_g$ denote the cost of sensor encoding, context encoding, and lightweight gating, respectively. 
A naive per-sample context-integration design incurs approximately $T(C_x+C_c+C_g)$ cost over a stream of $T$ samples. 
With caching, \framework{} reduces this to $T(C_x+C_g)+(M+1)C_c$, where $M \ll T$ in stable deployment periods. 
Thus, the expensive context branch is removed from the steady-state per-sample path, while the gated head still receives a context prior for every prediction.}

The cache stores the output of the learned context encoder after normalization, so that subsequent steps reuse the same context prior. 
This mechanism reduces both latency and energy consumption on resource-constrained edge devices without changing the downstream prediction behavior when the context remains unchanged.
The cache key $u_t$ can be obtained either from explicit deployment metadata or from a lightweight context detector. 
When the context detector is used, \framework{} does not require the detector to run at the same frequency as the sensing backbone.
Instead, the detector only needs to provide a context update when the deployment condition changes or when the cached identifier becomes stale.
This event-driven design is important for edge deployment: context processing is moved away from the per-sample inference path, while the prediction head still receives a context prior for every sensor segment through the cache.

\textbf{Context acquisition and detection.}
\label{sec:context_prediction}
\framework{} exploits a lightweight context signal at inference time, which can be obtained in two practical ways. 
The first is directly acquired context, utilizing deployment metadata, OS/device tags, or explicit user setup. 
For example, IMU on-body placements such as ``Wrist'' and ``Belt'' can be provided directly via wearable OS metadata or initialized through lightweight user setup~\cite{shoaib}. 
The second is detected context, produced by a lightweight background classifier running on commodity signals. 
For instance, in our speech and WiFi pipelines, ambient environments such as ``Bus'' or ``Office'' are opportunistically inferred from low-power smartphone signals, such as WiFi scans or coarse audio statistics, following the StudentLife sensing paradigm~\cite{wang2014studentlife}. 
In our system evaluations, this lightweight front-end classifier achieves an automated detection accuracy of 87.25\%, serving as a realistic operating point for edge deployments.

Regardless of the acquisition method, the resulting context is formatted as a short language descriptor and encoded by $f_{\text{context}}$. 
This design decouples context acquisition from the neural architecture, so that the downstream model consumes context in a uniform format.

\section{Evaluation}
\label{sec:evaluation}



\subsection{Methodology}

\subsubsection{Datasets}
We use three public datasets covering inertial sensing, speech, and WiFi signals (Table~\ref{tab:public_datasets_summary}). We use Shoaib with on-body placement as context, VoiceBank-DEMAND with noise types as acoustic context, and the FallDetection subset of CSI-Bench with environment as context. Table~\ref{tab:mmd_scenarios} summarizes the source--target context splits and latent MMD values used to construct the Low/Mid/High shift settings.

\begin{table}
    \centering
    \scriptsize
    \setlength{\tabcolsep}{2pt}
    \renewcommand{\arraystretch}{1.15}
    \begin{tabularx}{\columnwidth}{@{} l c >{\raggedright\arraybackslash}p{1.4cm} >{\raggedright\arraybackslash}p{1.72cm} c @{}}
        \toprule
        \textbf{Dataset} & \textbf{Modality} & \textbf{Task} & \textbf{Context (\#)} & \textbf{\# of subjects} \\
        \midrule
        Shoaib~\cite{shoaib} & IMU & HAR & Placement (5) & 10 \\
        VoiceBank-DEMAND~\cite{valentini2016investigating} & Speech & Enhancement & Environment (10) & 30 \\
        CSI-Bench~\cite{zhu2025csibench} & WiFi & Fall Detection & Environment (6) & 17 \\
        \bottomrule
    \end{tabularx}
    \caption{Summary of datasets used in the evaluation.}
    \label{tab:public_datasets_summary}
\end{table}

\subsubsection{Experimental Settings}
To study robustness under controlled context shifts, we construct Low/Mid/High tiers using latent-MMD distances between fixed source contexts and unseen target contexts (Table~\ref{tab:mmd_scenarios}). For each modality, we use two source contexts for training and one held-out target context for testing in each tier. We pre-train the effective context representation learning module on the 80\% training split of the source contexts to learn compact sensing-aligned context representations, and train the efficient context integration module with 1/2/5/10\% labeled source data from the same contexts. Unless otherwise noted, all results follow a strict source-only protocol without target-domain data; §\ref{sec:overall_performance} separately studies limited target-label adaptation.
\textcolor{blue}{For efficient context integration, we use a lightweight gating module with compact modality-appropriate cues, while relying on dynamic context caching to avoid repeated context encoding.}
\textcolor{blue}{This protocol separates representation learning, source-supervised integration, and target-context evaluation: unlabeled source sensor--context pairs learn compact sensing-aligned context representations, while labeled source samples train the efficient context integration module. 
The held-out target context is never used for training, hyperparameter selection, or model updates; during evaluation, only its context description is provided to \framework{}.}

\begin{table}[tbp]
  \centering
  \small
  \setlength{\tabcolsep}{3pt}
  \renewcommand{\arraystretch}{1.1}
  \begin{tabularx}{\columnwidth}{@{} l l X c r @{}}
    \toprule
    \textbf{Modality} & \textbf{Source contexts} & \textbf{Target} & \textbf{Shift} & \textbf{Latent MMD} \\
    \midrule
    \multirow{3}{*}{IMU} 
      & \multirow{3}{*}{left/right pocket} 
      & wrist      & Low  & 0.587 \\
      & & upper arm & Mid  & 0.873 \\
      & & belt      & High & 1.571 \\
          \midrule
    \multirow{3}{*}{Speech} 
      & \multirow{3}{*}{kitchen, restaurant} 
      & living room & Low  & 0.095 \\
      & & bus         & Mid  & 0.194 \\
      & & office      & High & 0.202 \\
        \midrule
    \multirow{3}{*}{WiFi} 
      & \multirow{3}{*}{large house, office} 
      & medium house & Low  & 0.067 \\
      & & apartment    & Mid  & 0.100 \\
      & & villa        & High & 0.214 \\

    \bottomrule
  \end{tabularx}
  \caption{Settings of context shifts for different datasets. }
  \label{tab:mmd_scenarios}
\end{table}

\begin{figure*}[tb]
    \centering
    \setlength{\abovecaptionskip}{0.cm}
    \setlength{\belowcaptionskip}{0.cm}
    \includegraphics[width=\textwidth]{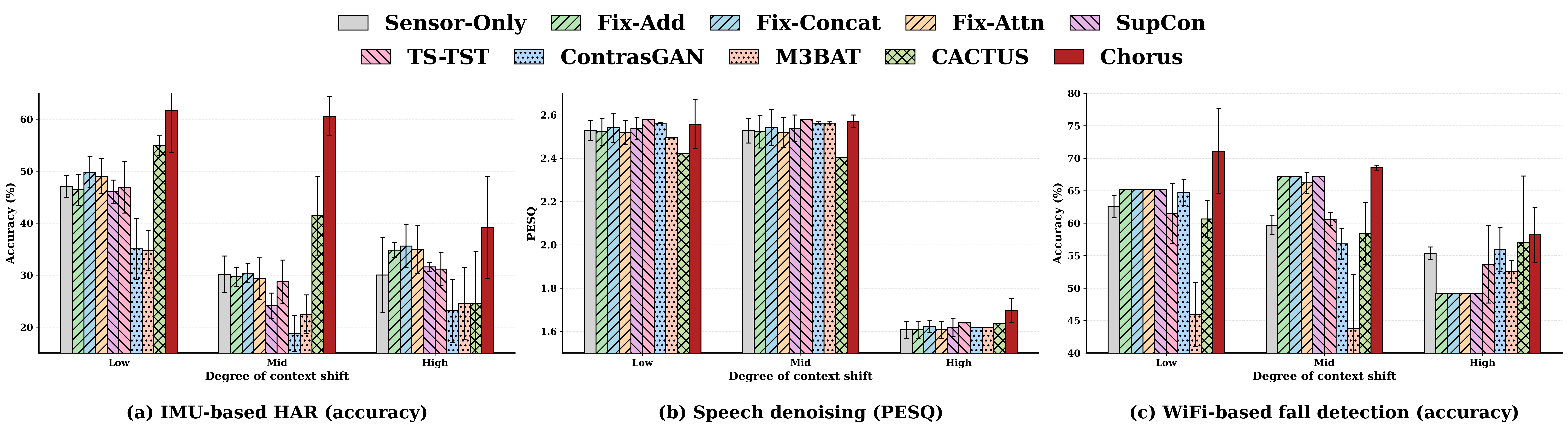}
    \caption{Performance across degrees of context shifts for IMU-based human activity recognition (HAR), speech enhancement, and WiFi-based fall detection. Error bars indicate standard deviation over three random seeds.}
    \label{fig:overall_difficulty_10pct}
\end{figure*}

\subsubsection{Baselines}
We compare \framework{} with representative baselines from several families of sensing adaptation and context-integration methods, covering source-only sensing, fixed context fusion, representation learning, domain adaptation/generalization, and context-specialist designs:

\begin{itemize}
    \item \textbf{Sensor-only baseline.} 
    \textit{Sensor-Only} uses the same sensor backbone as \framework{} but removes all context inputs. 
    This baseline measures how well a standard source-trained sensing model transfers to unseen contexts without any contextual prior.

    \item \textbf{Fixed context fusion baselines.} 
    \textit{Fix-Add}, \textit{Fix-Concat}, and \textit{Fix-Attn} inject the context embedding into the sensor representation through addition, concatenation, or attention-based fusion, respectively. 
    These baselines evaluate whether simple context injection is sufficient for unseen-context customization.

    \item \textbf{Representation learning baselines.} 
    \textit{SupCon} applies supervised contrastive learning to improve feature separability under the same source-label budget. 
    \textit{TS-TST}~\cite{tst} is a source-only time-series pre-training/fine-tuning baseline that improves temporal feature extraction without explicitly modeling deployment context.

    \item \textbf{Domain adaptation and generalization baselines.} 
    \textit{M3BAT}~\cite{meegahapola2024m3bat} and \textit{ContrasGAN}~\cite{ContrasGAN} represent domain-adaptation-style and contrastive/adversarial approaches for robust sensing under distribution shifts. 
    Since our main setting does not allow target-domain sensing samples, we evaluate their source-only variants for fair comparison.

    \item \textbf{Context-specialist baseline.} 
    \textit{CACTUS}~\cite{CACTUS} is a context-aware IoT inference system that dynamically switches among context-specific micro-classifiers. 
    It represents specialist-style context modeling, where model structure grows with context-specific experts.
\end{itemize}

Unless otherwise noted, all baselines are trained under the same source-only protocol, source-label budgets, and backbone capacity where applicable. 
For methods originally designed for target-aware adaptation, we report their source-only variants to match the data-free customization setting of \framework{}.

\subsubsection{Configurations} 
Within each modality, we use a compact 1D CNN backbone for IMU HAR and a Transformer encoder for speech enhancement and WiFi fall detection, with shared hidden dimensions and depth across methods.\footnote{TS-TST~\cite{tst} is the only baseline that requires its own Transformer-based architecture.} 
\textcolor{blue}{For \framework{}, the effective context representation learning module is pre-trained on the 80\% source-context split to learn compact sensing-aligned context representations, and the efficient context integration module is trained on 1/2/5/10\% labeled source data.} Representation pre-training and source-supervised integration training both run for up to 100 epochs with patience 15. 
Batch size and learning rate are fixed per modality across all source-budget settings (IMU: 32 and $10^{-4}$; WiFi: 64 and $5\times10^{-5}$; Speech: 2 and $5\times10^{-5}$). 
We report accuracy for IMU/WiFi classification tasks and PESQ for speech enhancement. 
All reported results are averaged over three random seeds, and error bars indicate standard deviation unless otherwise specified. 
Model selection is performed using source validation data only, and no target-context samples are used for early stopping or hyperparameter tuning.

\subsubsection{Implementation}
We implement all learning-based methods in the same training pipeline and deploy \framework{} on two commodity smartphones for IMU HAR: an iPhone 16 Pro running CoreML and a Xiaomi 14 running ONNX Runtime Mobile. 
On both platforms, inference runs on CPU for controlled comparison, with simple sensor-wise IMU calibration before inference. 
The mobile deployment is used to measure inference latency, memory footprint, and energy overhead, while the main accuracy comparisons are conducted under the controlled source-only evaluation protocol described above.

\subsection{Overall Performance}
\label{sec:overall_performance}

\textcolor{blue}{We compare \framework{} and the baselines across modalities and context-shift levels under the data-free post-deployment customization setting, where no target-domain sensing samples or post-deployment model updates are used.} Figure~\ref{fig:overall_difficulty_10pct} fixes the labeled source budget to 10\% and varies the degree of context shift, while Figure~\ref{fig:story_source_target} focuses on the hardest IMU target (\textit{Belt}) under different source-label budgets and limited target-label adaptation.

\subsubsection{Performance across different degrees of context shift.}
With 10\% labeled source data, \framework{} remains strong as context shift increases. On IMU and WiFi, it outperforms the strongest baseline across all three shift tiers, with the clearest margin on IMU Low shift, where it improves over CACTUS by 20.2\%. Generic adversarial and contrastive baselines can remain competitive under lighter shifts, but degrade more noticeably as the shift becomes stronger.
\textcolor{blue}{On speech enhancement, \framework{} remains competitive in the Low and Mid tiers and achieves the best result under the High shift, where source-only transfer becomes less reliable. Overall, these results show that compact sensing-aligned context representations provide consistent benefits under unseen context shifts, especially when fixed fusion and generic source-only baselines degrade.}

\subsubsection{Performance with different amounts of data.}

\begin{figure}
    \centering
    \setlength{\abovecaptionskip}{0.cm}
    \setlength{\belowcaptionskip}{0.cm}
    \includegraphics[width=\columnwidth]{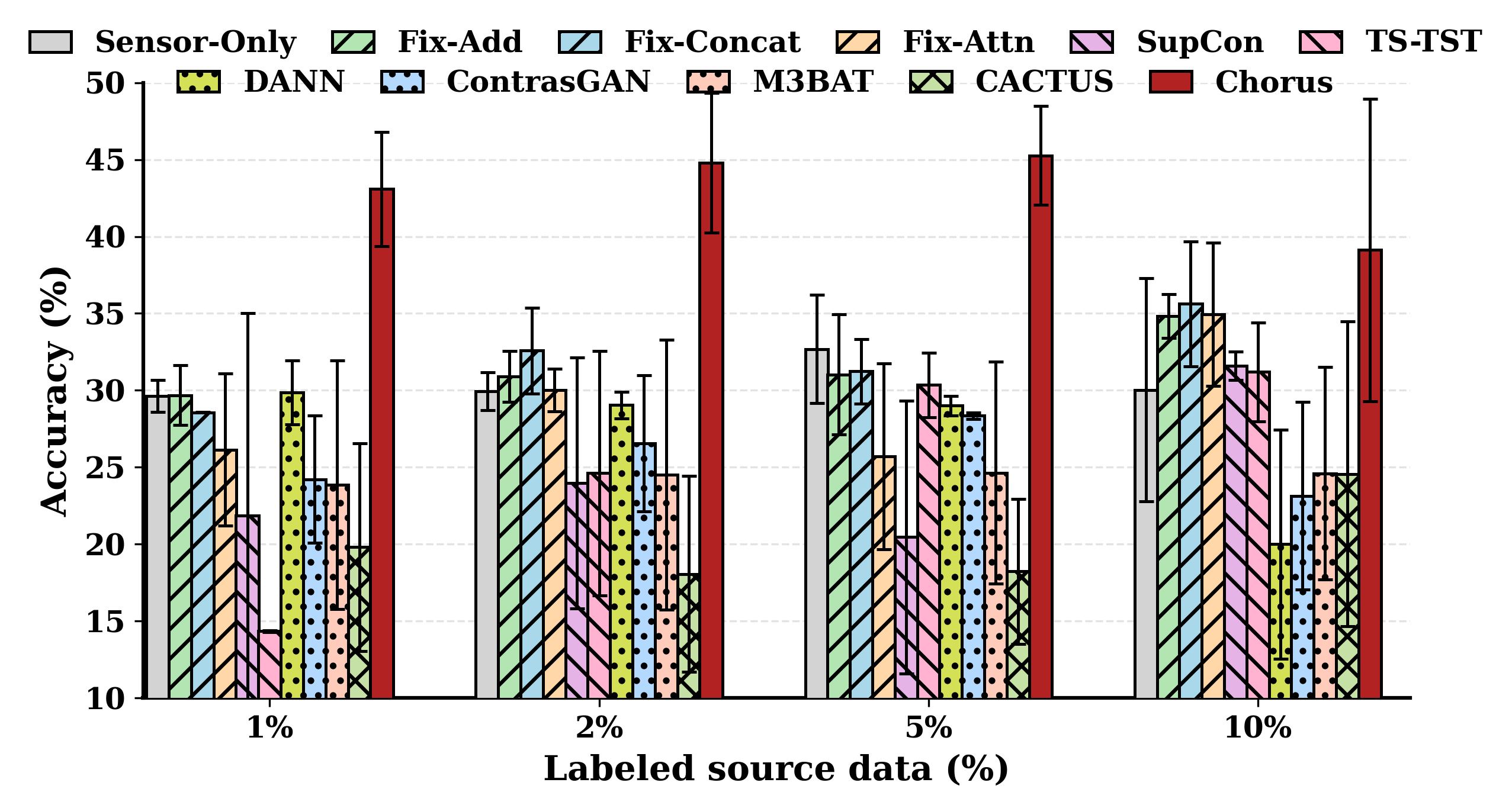}
    \caption{Performance with different amounts of labeled data. Error bars indicate standard deviation over three random seeds.}
    \label{fig:story_source_target}
\end{figure}

We next vary the labeled source-data budget on the hardest IMU target (\textit{Belt}). As shown in Figure~\ref{fig:story_source_target}(a), \framework{} remains the strongest method across all budgets, but its performance is not monotonic: it improves from 1\% to 5\% and then drops at 10\%, where the cross-seed variance is also the largest. This suggests that under severe unseen shift, more labeled source data does not necessarily improve target-context generalization and may instead increase over-specialization to source-context-specific patterns. 

\subsubsection{Few-shot test-time adaptation under the hardest shift.}
\begin{figure}
    \centering
    \setlength{\abovecaptionskip}{0.cm}
    \setlength{\belowcaptionskip}{0.cm}
    \begin{subfigure}[t]{0.47\columnwidth}
        \centering
            \setlength{\abovecaptionskip}{0.cm}
    \setlength{\belowcaptionskip}{0.cm}
        \includegraphics[width=\linewidth]{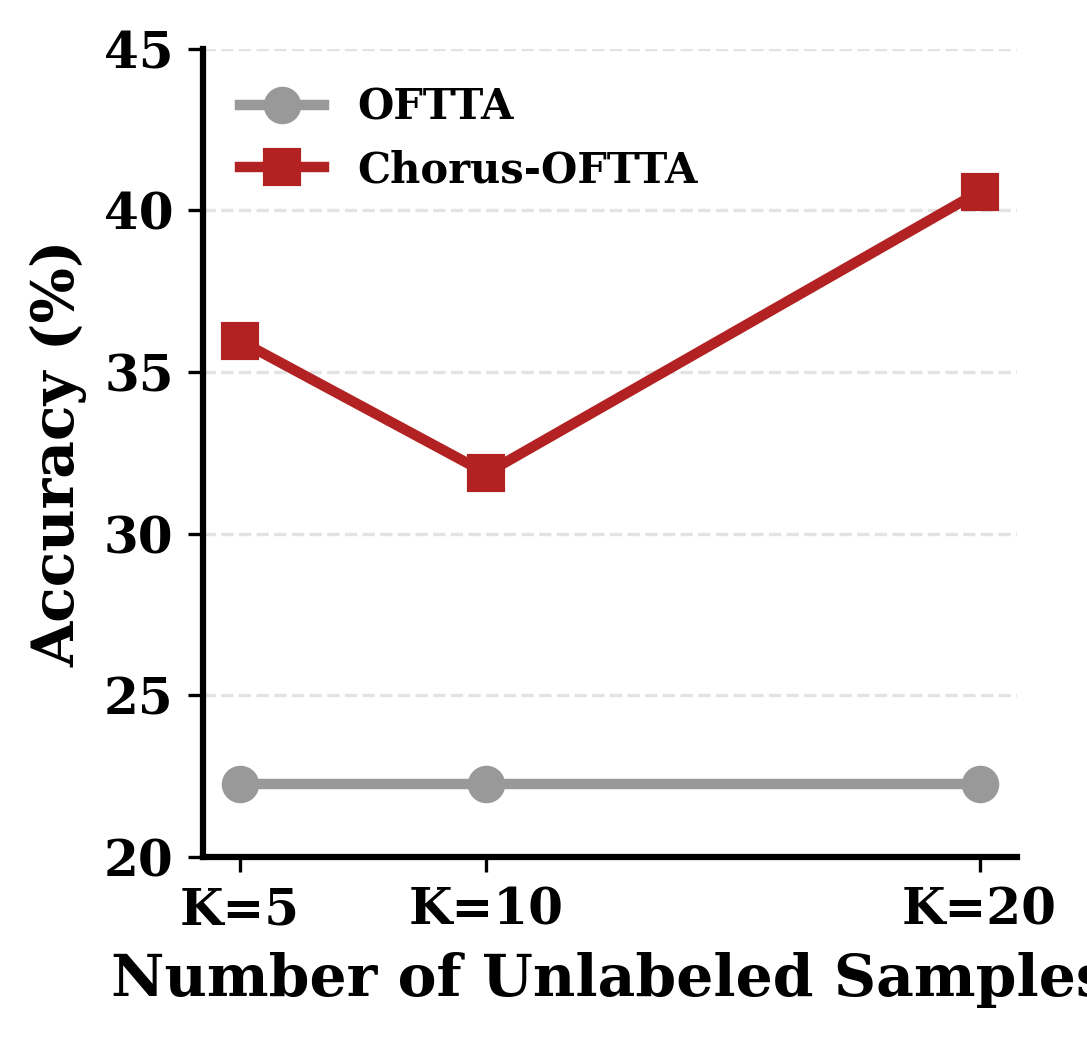}
        \caption{Unlabeled target samples.}
        \label{fig:fewshot_unlabeled}
    \end{subfigure}
    \hfill
    \begin{subfigure}[t]{0.47\columnwidth}
        \centering
            \setlength{\abovecaptionskip}{0.cm}
    \setlength{\belowcaptionskip}{0.cm}
        \includegraphics[width=\linewidth]{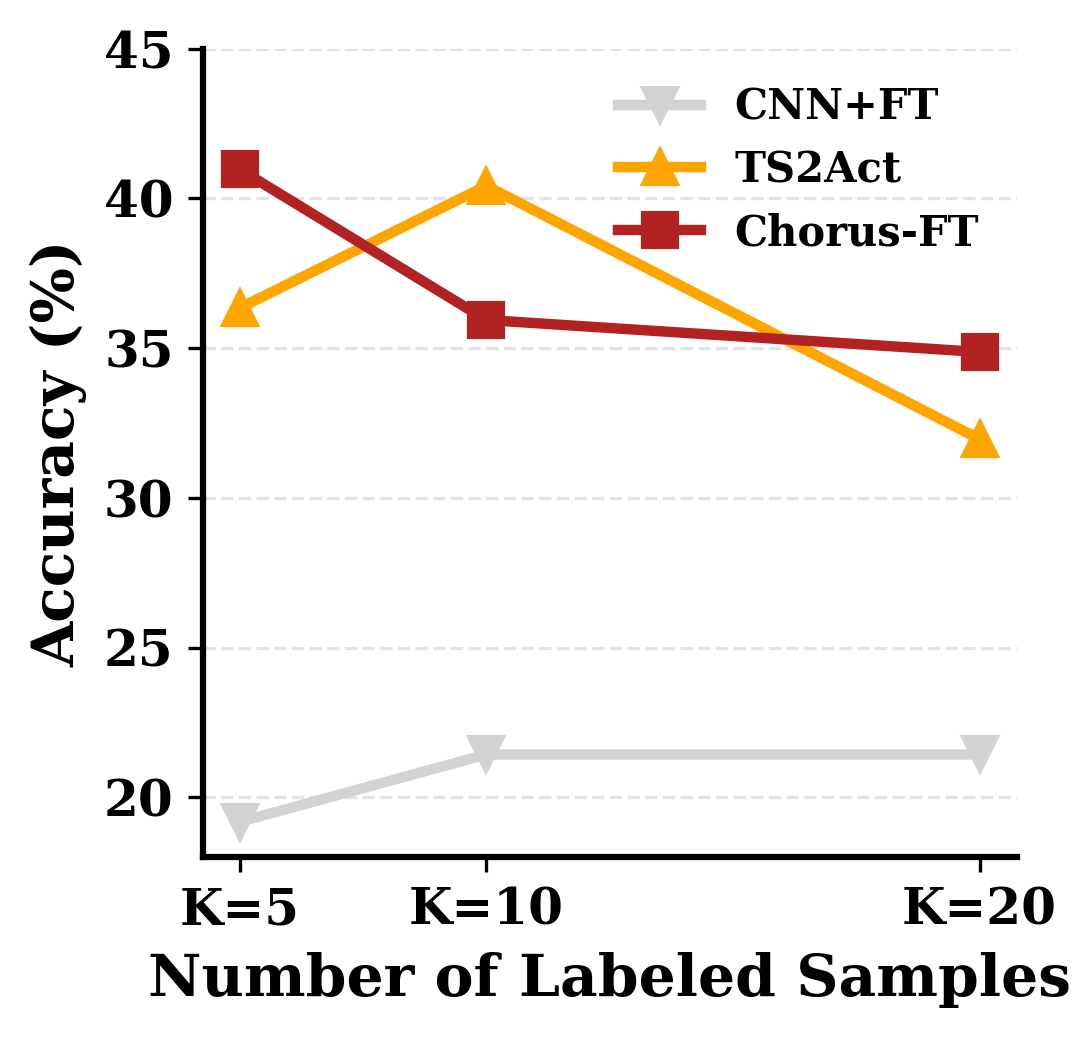}
        \caption{Labeled target samples.}
        \label{fig:fewshot_labeled}
    \end{subfigure}
    \caption{Few-shot adaptation on the hardest IMU shift (\textit{Belt}).}
    \label{fig:fewshot_target_adaptation}
\end{figure}

As an additional diagnostic beyond the main data-free setting, we examine the same source-trained \framework{} model when a small amount of target data becomes available under the hardest IMU setting (High shift, \textit{Belt}). At test time, we expose the model to only $K\in\{5,10,20\}$ target samples per class under two settings: unlabeled target samples and labeled target samples. \framework{}-OFTTA applies an OFTTA-style unlabeled adaptation protocol, while \framework{}-FT performs supervised few-shot adaptation from the same source-trained initialization.
Figure~\ref{fig:fewshot_target_adaptation} shows that unlabeled few-shot adaptation remains unstable and non-monotonic under this shift, indicating sensitivity to the few-shot target set. This suggests that post-hoc unlabeled adaptation alone is insufficient when the source-target geometry remains substantially mismatched.

\begin{figure}
    \centering
    \setlength{\abovecaptionskip}{0.cm}
    \setlength{\belowcaptionskip}{0.cm}

    \begin{subfigure}[t]{0.49\linewidth}
        \centering
            \setlength{\abovecaptionskip}{0.cm}
    \setlength{\belowcaptionskip}{0.cm}
        \includegraphics[width=\linewidth]{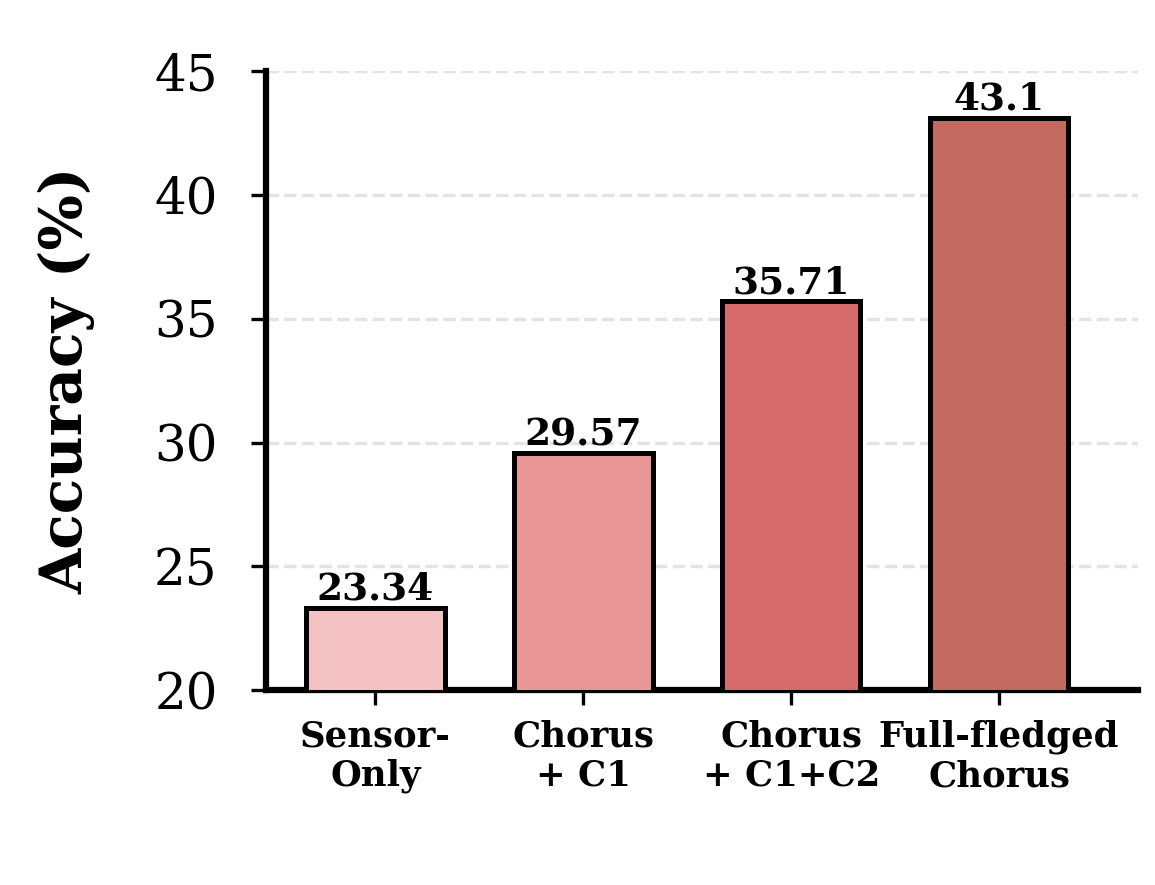}
        \subcaption{Design modules.}
        \label{fig:ablation_arch}
    \end{subfigure}
    \hfill
    \begin{subfigure}[t]{0.49\linewidth}
        \centering
            \setlength{\abovecaptionskip}{0.cm}
    \setlength{\belowcaptionskip}{0.cm}
        \includegraphics[width=\linewidth]{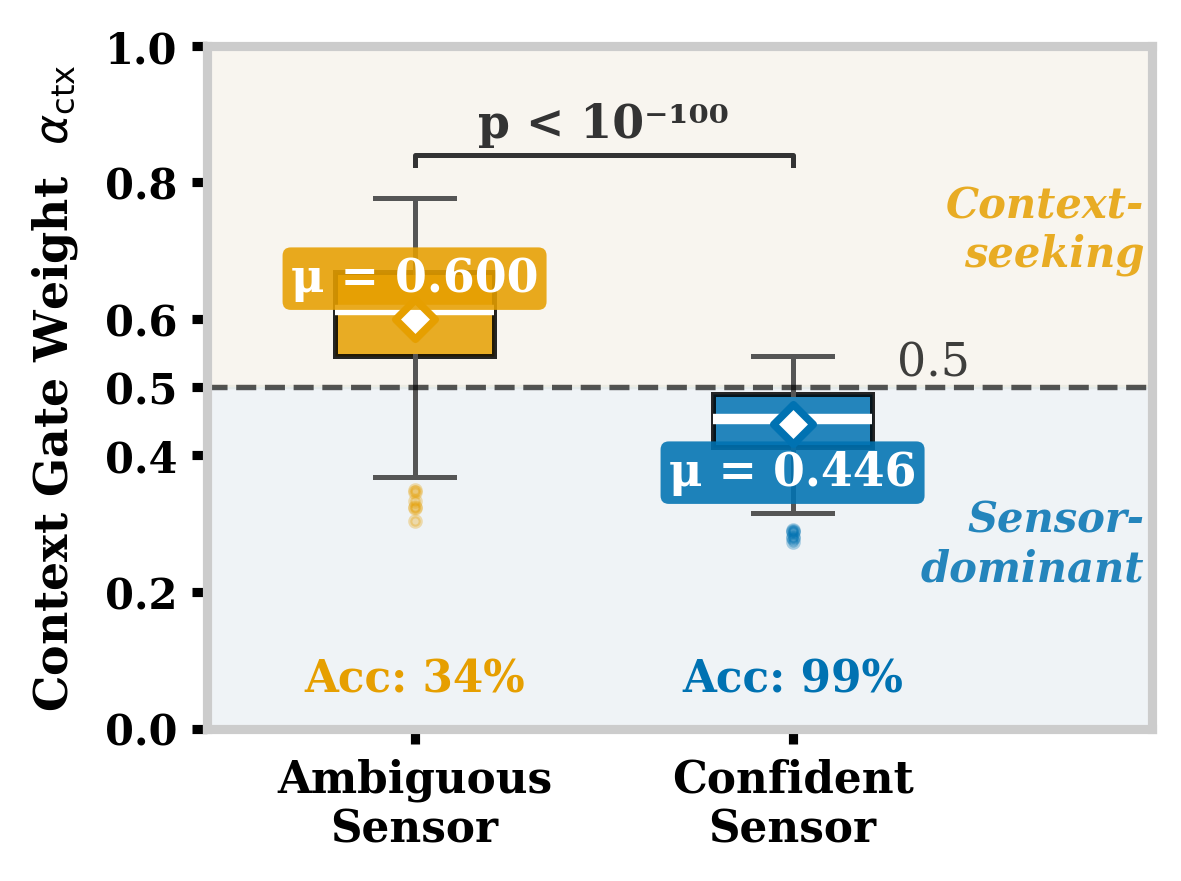}
        \subcaption{Sample-level gating analysis.}
        \label{fig:gating_boxplot}
    \end{subfigure}
    \hfill
    \begin{subfigure}[t]{\linewidth}
        \centering
        \includegraphics[width=\linewidth]{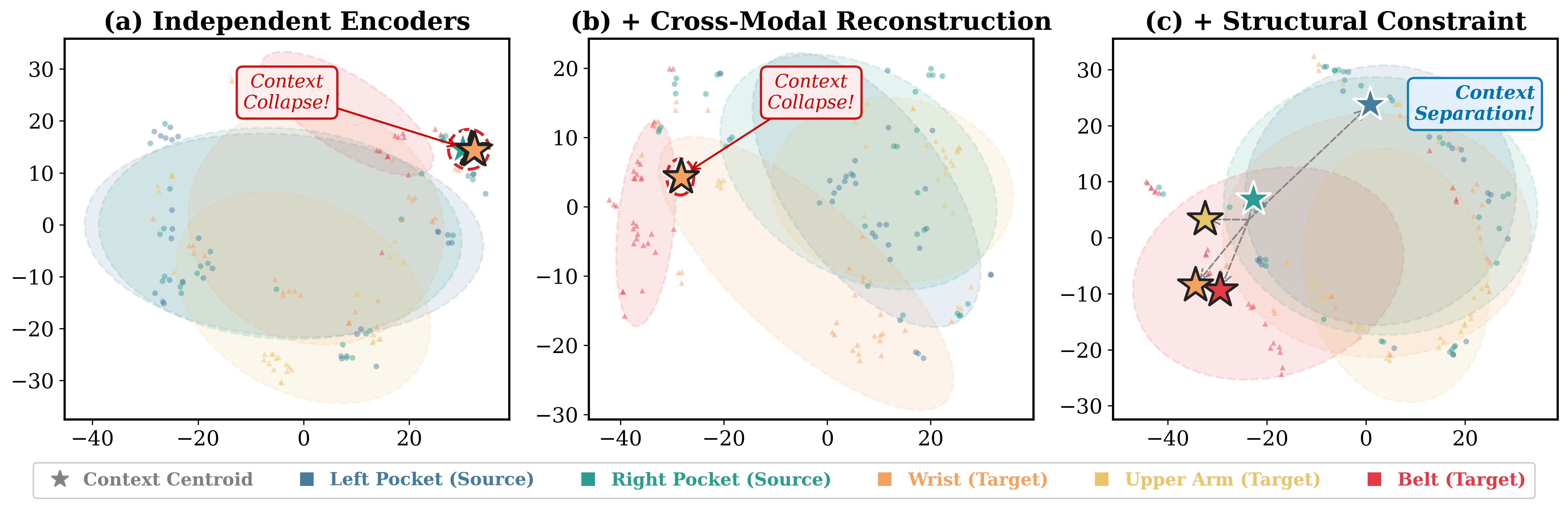}
        \subcaption{t-SNE~\cite{vandermaaten2008tsne} visualization of source/target sensor features and context centroids. Star markers denote context centroids.}
        \label{fig:tsne_latent_imu}
    \end{subfigure}

    \caption{Understanding \framework{}'s performance.}
    \label{fig:ablation_combined}
\end{figure}

\subsection{Understanding \framework{}'s performance}
\label{sec:ablation_study}
We next examine which components of \framework{} drive robustness under context shifts.

\subsubsection{Ablation study.}
\textcolor{blue}{Effective context representation learning is the main source of robustness, while efficient context integration provides additional inference-time gains. On the hardest IMU setting (1\% labels, High shift), static context fusion improves the Sensor-Only backbone from 23.3\% to 29.6\%, and lightweight gating further raises it to 35.7\%. With the context representation module enabled, the full \framework{} reaches 43.1\%. This indicates that integration helps, but the dominant gain comes from learning compact, transferable, and sensing-aligned context representations that better capture context-induced sensor variation.}

\subsubsection{Effectiveness of context representations.}
\textcolor{blue}{\framework{} improves robustness not by simply injecting context, but by learning compact, transferable, and sensing-aligned context representations under unseen shifts. Without context-space regularization, the learned representations still suffer from context collapse, with centroid cosine similarity increasing to 0.727 and mean pairwise L2 distance shrinking to 3.65. After adding KL compactness regularization and context-identity contrastive separation, centroid cosine similarity drops to $-0.221$ and mean pairwise L2 distance expands to 12.94, yielding much clearer placement-specific clusters in Figure~\ref{fig:tsne_latent_imu}. Star markers denote context centroids.}

\subsubsection{Effectiveness of the gating module.}
\textcolor{blue}{The lightweight gating module operates at the sample level rather than following a fixed environment-level policy. As shown in Figure~\ref{fig:gating_boxplot}, it assigns lower context weight to high-confidence sensor samples and higher context weight to ambiguous ones. This suggests that the lightweight gating module is an effective part of efficient context integration, using the learned context prior when it is useful for the current sensor segment.}

\subsection{Robustness to Context Inputs}
\label{sec:predicted_context_robustness}

\subsubsection{Robustness to imperfect context prediction}

\begin{figure}
    \centering
    \setlength{\abovecaptionskip}{0cm}
    \setlength{\belowcaptionskip}{0cm}

    \begin{subfigure}{0.32\columnwidth}
        \centering
        \setlength{\abovecaptionskip}{0cm}
        \setlength{\belowcaptionskip}{0cm}
        \includegraphics[width=\linewidth]{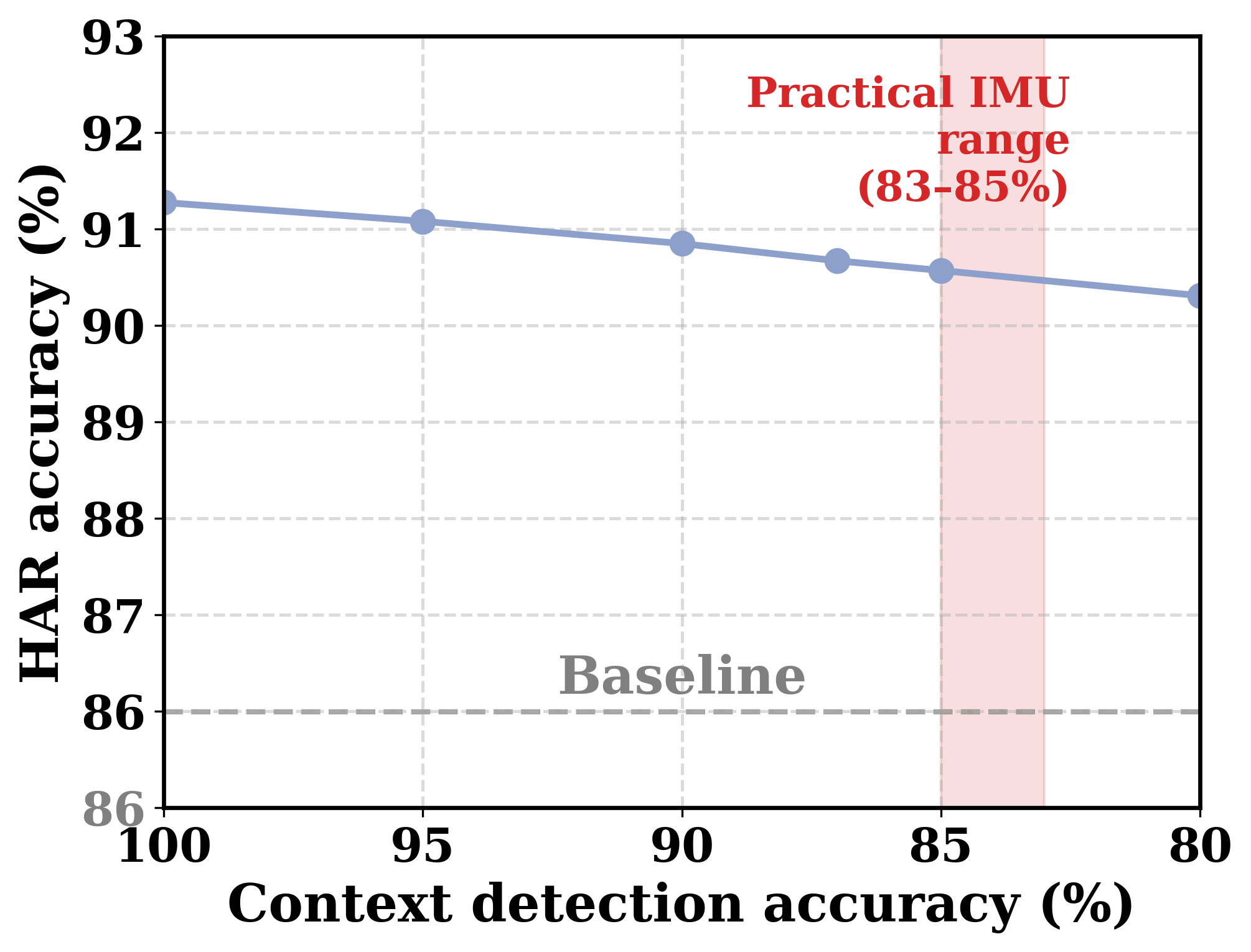}
        \caption{IMU.}
        \label{fig:ctx_robust_imu}
    \end{subfigure}
    \begin{subfigure}{0.32\columnwidth}
        \centering
        \setlength{\abovecaptionskip}{0cm}
        \setlength{\belowcaptionskip}{0cm}
        \includegraphics[width=\linewidth]{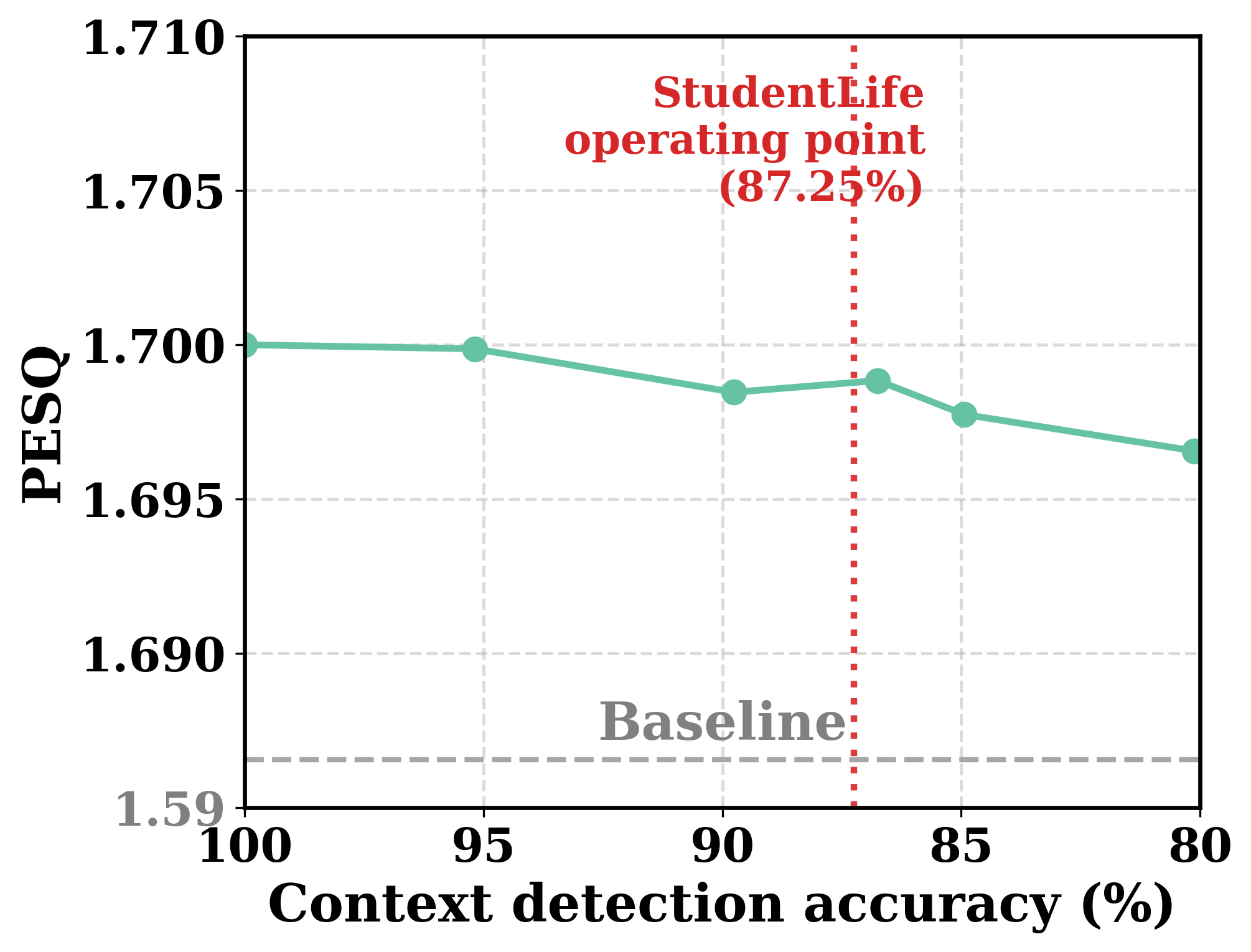}
        \caption{Speech.}
        \label{fig:ctx_robust_speech}
    \end{subfigure}
    \begin{subfigure}{0.32\columnwidth}
        \centering
        \setlength{\abovecaptionskip}{0cm}
        \setlength{\belowcaptionskip}{0cm}
        \includegraphics[width=\linewidth]{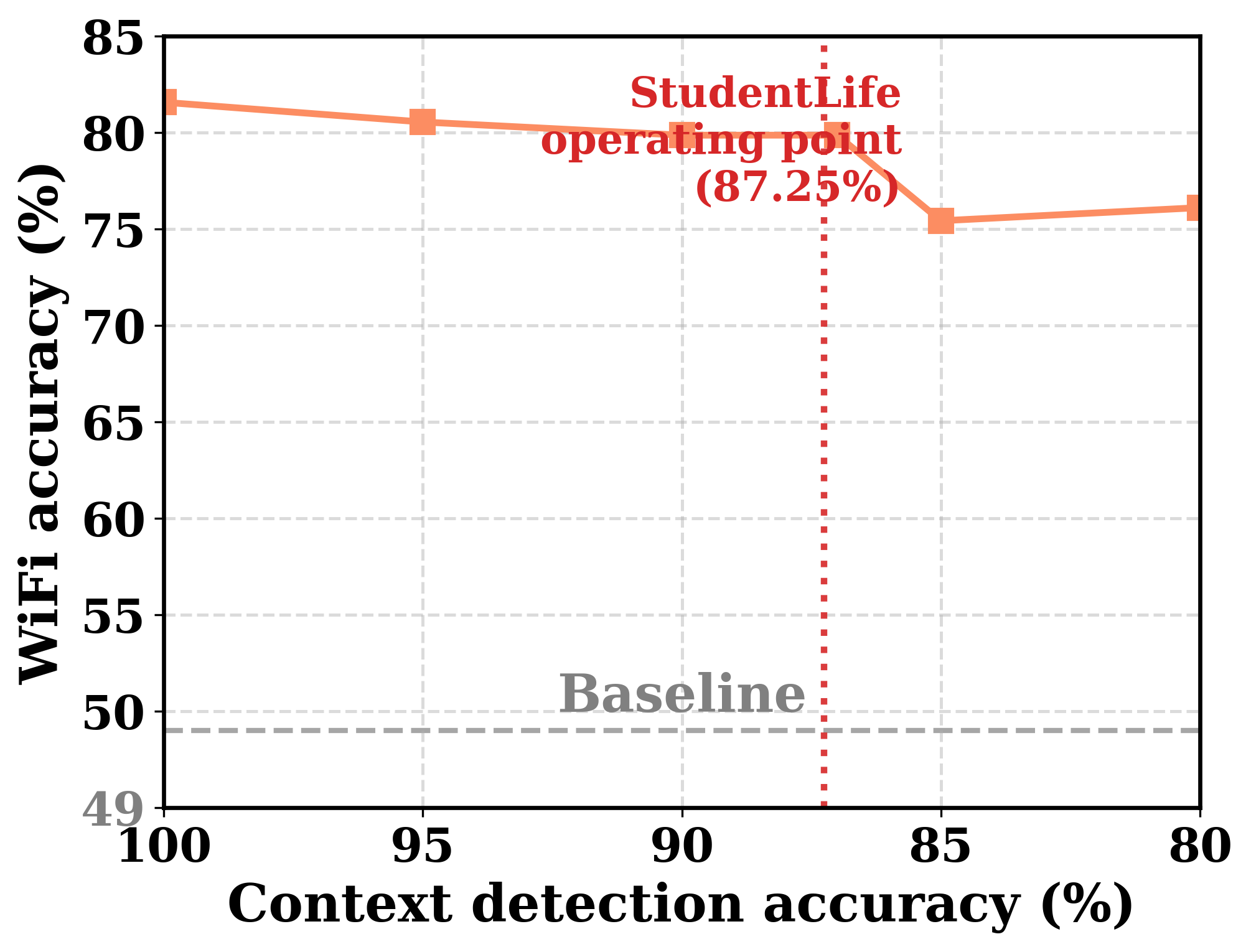}
        \caption{WiFi.}
        \label{fig:ctx_robust_wifi}
    \end{subfigure}

    \caption{Robustness to imperfect context prediction. The horizontal dashed line in each panel marks the corresponding Sensor-Only baseline, while the red marker indicates a realistic deployment operating point.}

    \label{fig:context_robustness}
\end{figure}

\textbf{Context detection operating points.} In deployment, context is typically provided by a lightweight front-end rather than oracle metadata. To evaluate this setting, we use the 10\% labeled, High-shift checkpoints of \framework{} and vary only the accuracy of the upstream context detector at inference time. For speech and WiFi, we use the StudentLife-style operating point of 87.25\% context-detection accuracy. For IMU placement, a lightweight setup signal over a short initialization window can provide a practical deployment range of roughly 83--85\% context-detection accuracy. These operating points are marked in Figure~\ref{fig:context_robustness} as realistic deployment references rather than oracle settings.

\textbf{Impact on downstream performance}
Figure~\ref{fig:context_robustness} shows how downstream performance changes as context-detection accuracy decreases. In each panel, the horizontal dashed line marks the corresponding Sensor-Only baseline, i.e., 30.0 for IMU, 1.59 for speech, and 49.0 for WiFi. \textcolor{blue}{Across all three modalities, \framework{} remains above this baseline throughout the tested range, indicating that imperfect predicted context weakens the context prior but does not remove the benefit of context-bridged customization.}
The gap remains clear at realistic deployment points. In IMU, performance within the practical 83--85\% placement-detection range stays well above the Sensor-Only baseline. In speech and WiFi, the StudentLife operating point of 87.25\% also preserves a clear margin over the baseline, and performance degrades only gradually as context accuracy decreases further. This trend suggests that the effect of context noise is smooth rather than catastrophic: even when the predicted context is imperfect, the learned sensor representation still preserves substantial downstream task information.

\begin{figure}
    \centering
    \setlength{\abovecaptionskip}{0.cm}
    \setlength{\belowcaptionskip}{0.cm}
    \includegraphics[width=\columnwidth]{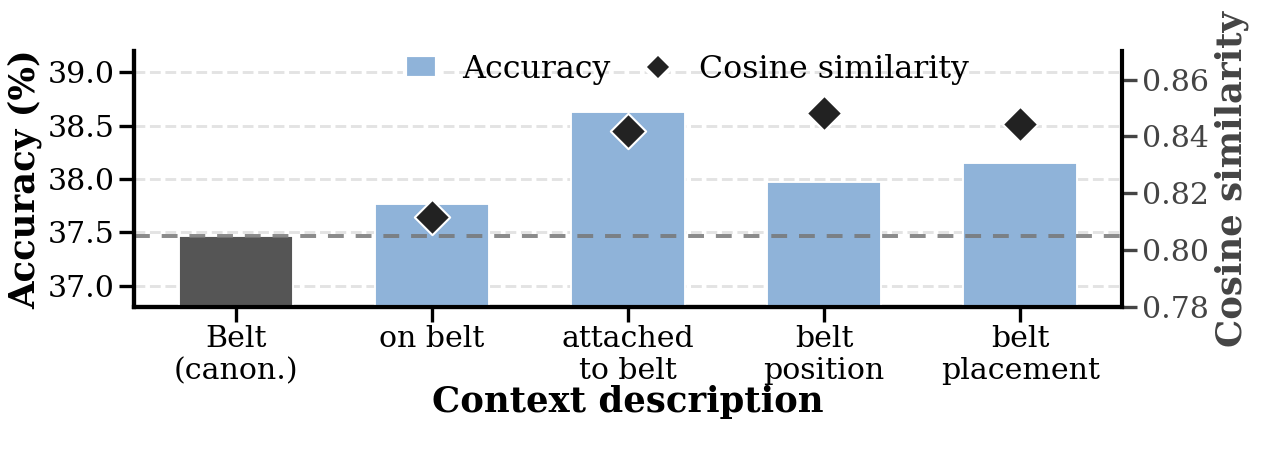}
    \caption{Robustness to different paraphrased context descriptions on the IMU High-shift context (\textit{Belt}). Bars show downstream accuracy under each context description, and diamond markers show cosine similarity between each paraphrased context representation and the canonical \textit{Belt} context representation. The dashed horizontal line marks the canonical-context accuracy.}
    \label{fig:context_description_variants}
\end{figure}

\subsubsection{Robustness to context description variants}

\textcolor{blue}{We further test whether \framework{} is sensitive to common wording variations of the same deployment context.
On the IMU High-shift target (\textit{Belt}), we keep the checkpoint, test split, labels, and target context ID fixed, and only replace the inference-time context description with paraphrased variants.
The canonical baseline uses the original \textit{Belt} context representation, while the paraphrased variants include \textit{on belt}, \textit{attached to belt}, \textit{belt position}, and \textit{belt placement}.}
\textcolor{blue}{As shown in Figure~\ref{fig:context_description_variants}, \framework{} remains stable under these paraphrased descriptions.
All paraphrased variants achieve accuracy above the canonical \textit{Belt} setting, with accuracy remaining in a narrow range around 38\%.
The cosine similarities between paraphrased and canonical context representations are also consistently high, indicating that the learned context space maps semantically equivalent descriptions to nearby representations.
This suggests that the learned compact sensing-aligned context space is robust to common wording variation, supporting effective context representation learning under practical deployment-time context descriptions.}

\subsection{Micro-benchmark Performance}
\label{sec:microbenchmark}
\begin{figure}[t]
    \centering
    \setlength{\abovecaptionskip}{0.cm}
    \setlength{\belowcaptionskip}{0.cm}
    \begin{subfigure}{0.32\columnwidth}
        \centering
        \includegraphics[width=\linewidth]{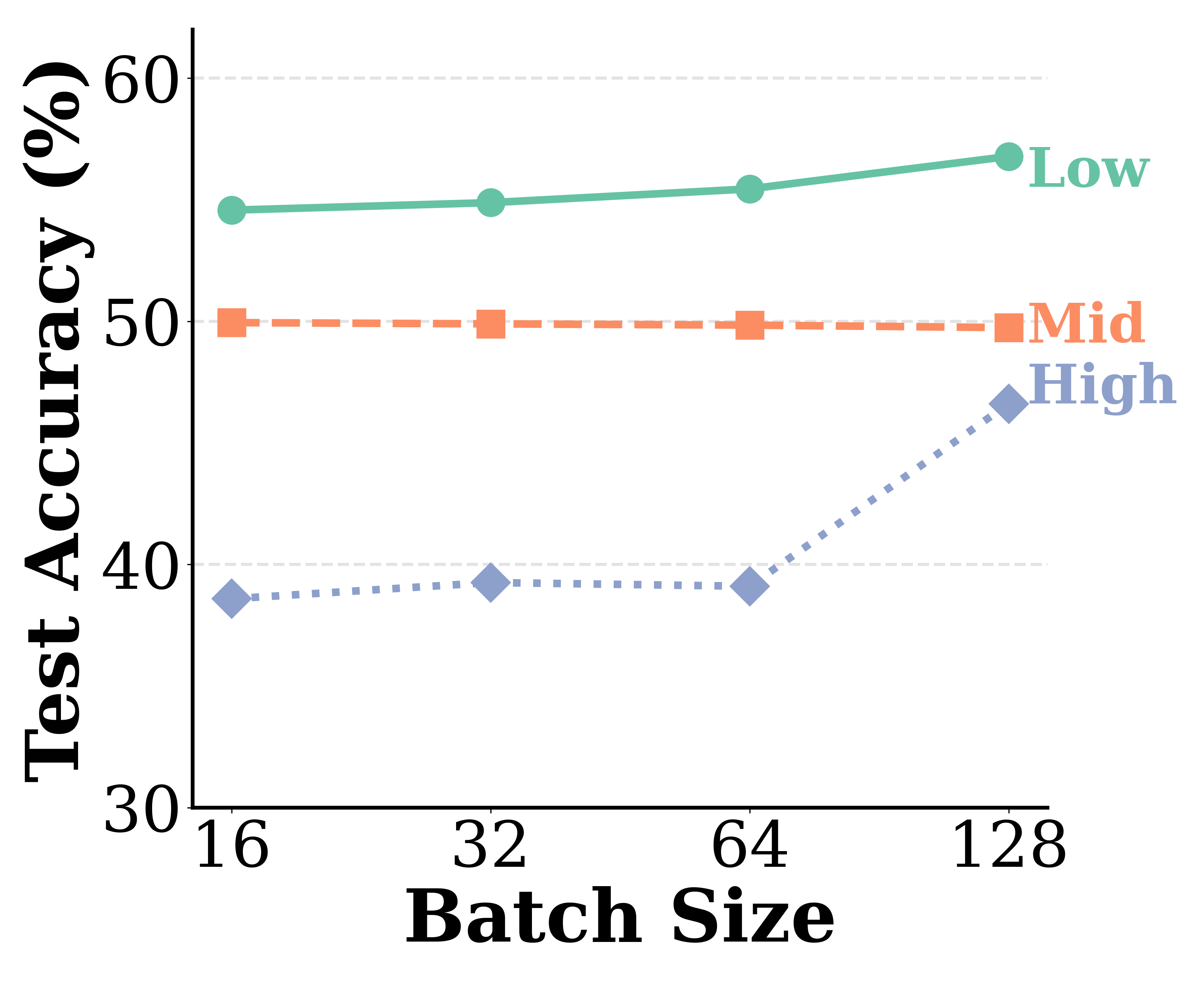}
        \subcaption{Batch size.}
        \label{fig:micro_batch}
    \end{subfigure}
    \hfill
    \begin{subfigure}{0.32\columnwidth}
        \centering
        \includegraphics[width=\linewidth]{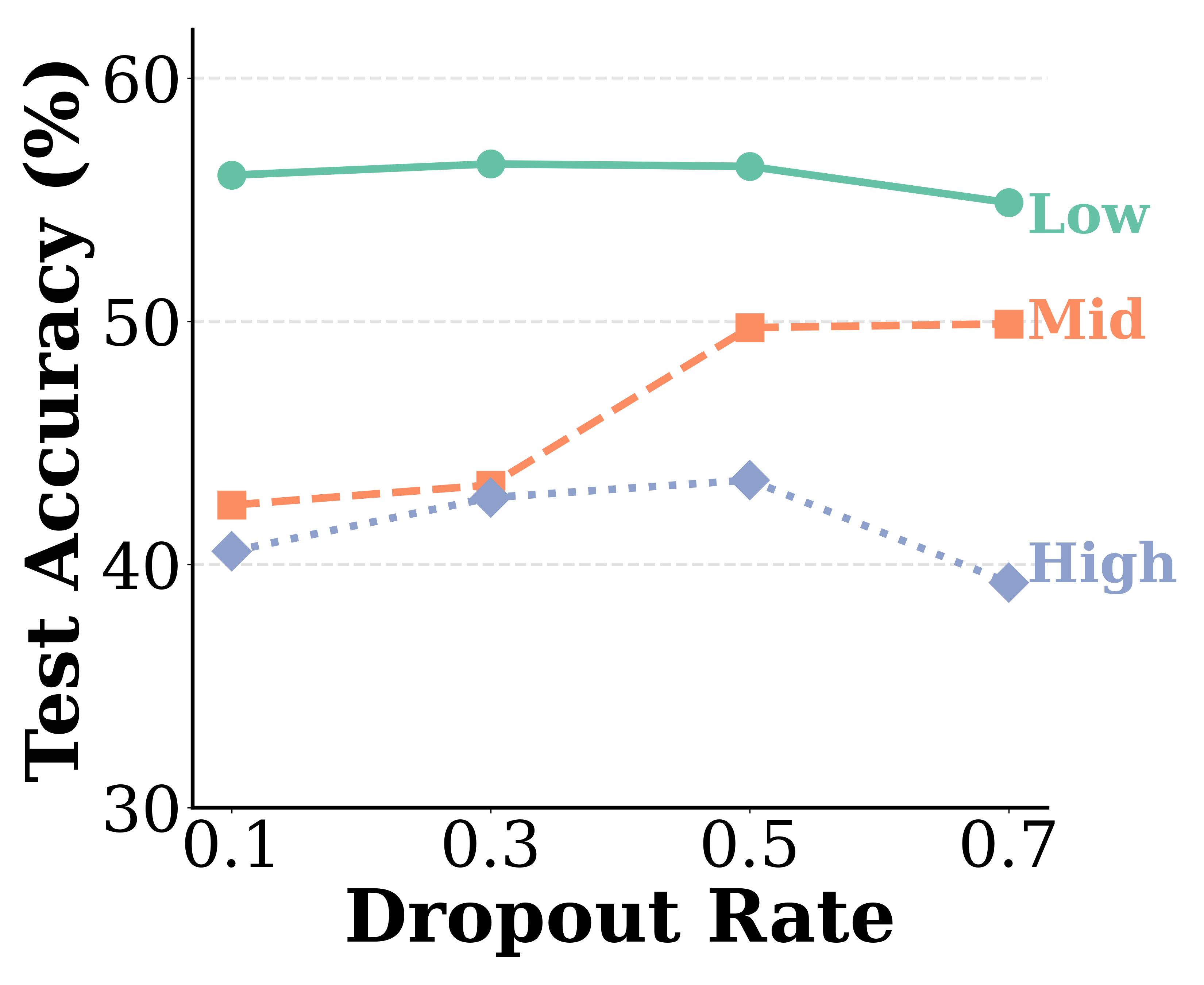}
        \subcaption{Dropout rate.}
        \label{fig:micro_dropout}
    \end{subfigure}
    \hfill
    \begin{subfigure}{0.32\columnwidth}
        \centering
        \includegraphics[width=\linewidth]{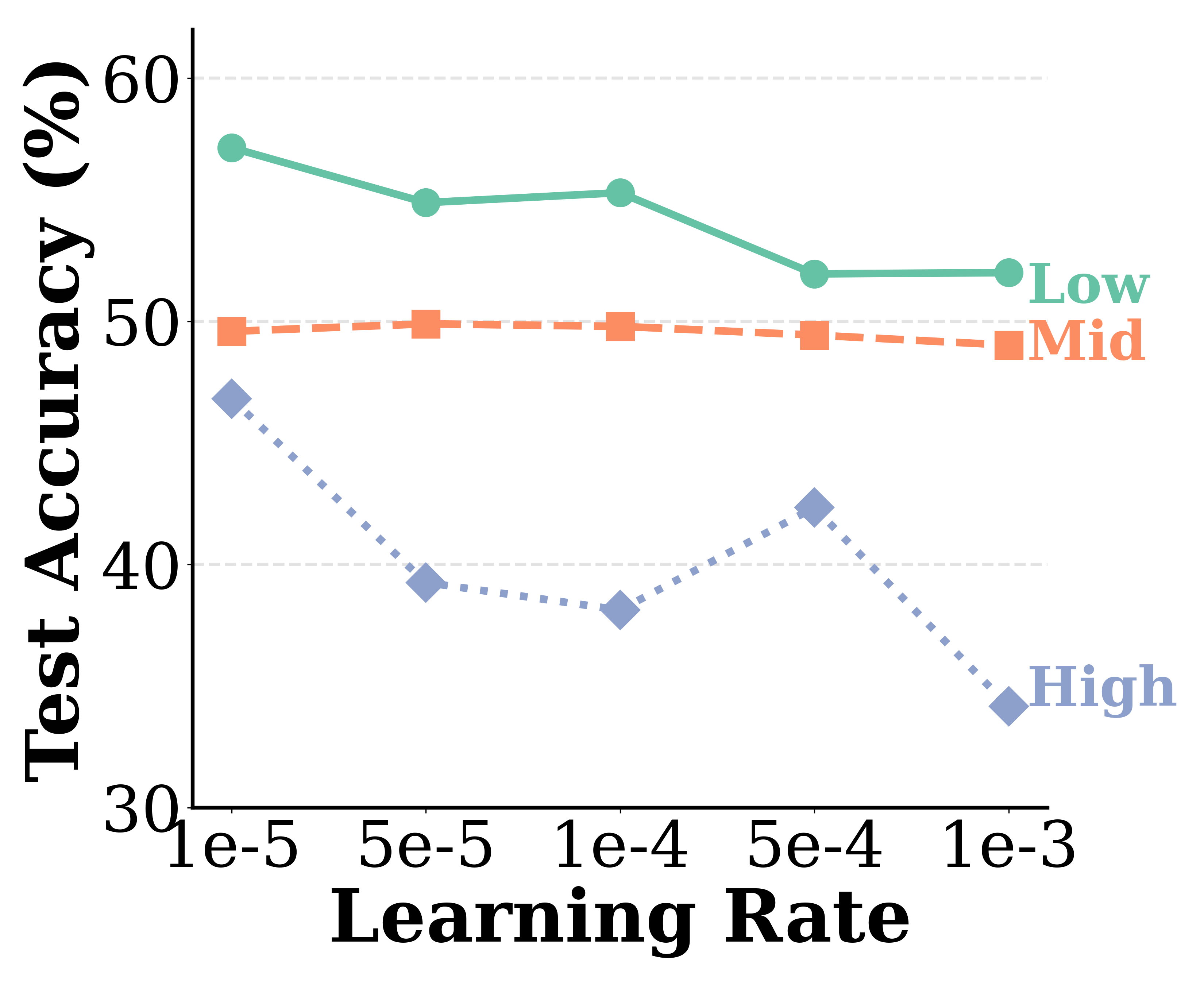}
        \subcaption{Learning rate.}
        \label{fig:micro_lr}
    \end{subfigure}
    \caption{Performance of \framework{} with different hyper-parameters.}
    \label{fig:microbenchmark}
\end{figure}

We next analyze the sensitivity of \framework{} to basic training hyperparameters in the IMU setting (Figure~\ref{fig:microbenchmark}).

\subsubsection{Impact of batch sizes.}
We first evaluate the accuracy with different batch sizes. The results are shown in Figure~\ref{fig:micro_batch}. We observe that the Low and Mid tiers remain almost unchanged from batch size 16 to 128, while the High tier improves noticeably at batch size 128. This means that \framework{} is generally robust to batch size, although the hardest-shift regime benefits more from a larger batch.

\subsubsection{Impact of dropout rates.}
We then study the impact of dropout rate on accuracy performance. The results are shown in Figure~\ref{fig:micro_dropout}. We observe that the Low tier stays stable across different dropout settings, while the Mid and High tiers vary more visibly and perform best under moderate dropout. This means that dropout mainly affects the more challenging shift regimes, where it serves as a useful regularizer for representation learning.

\subsubsection{Impact of learning rates.}
We further investigate the performance of \framework{} with different learning rates. The results are shown in Figure~\ref{fig:micro_lr}. We observe that the clearest sensitivity appears in the High tier, which performs best under the smallest learning rate and degrades under larger rates, while the Low and Mid tiers remain comparatively stable. This means that \framework{} is reasonably robust across a broad range of learning rates, but the hardest-shift regime prefers a more conservative optimization setting.

\subsection{System Overhead}
Beyond customization accuracy, efficient context integration should keep on-device overhead low. We therefore evaluate IMU inference latency, memory footprint, and smartphone-side energy on commodity mobile devices. Compared with CACTUS, which uses a context-specialist design, \framework{} maintains a unified generalist architecture and uses dynamic context caching to avoid repeated context encoding during streaming inference.

\subsubsection{Comparison of system overhead with baselines}
Table~\ref{tab:model_comparison} reports Xiaomi 14 overhead for IMU inference. With caching enabled, \framework{} runs at 0.547\,ms latency, remaining close to lightweight baselines and below CACTUS (0.648\,ms). Without caching, latency rises to 16.237\,ms, indicating that the dominant overhead comes from repeated context encoding. Although \framework{} keeps the context encoder loaded at runtime, it still uses substantially less memory than CACTUS on Xiaomi 14 (470.0\,MB vs.\ 854.27\,MB), because CACTUS maintains one micro classifier per context. Energy shows the same within-method trend, with caching reducing \framework{} from 4201.6\,mJ to 3287.3\,mJ.

\subsubsection{Impact of context caching across mobile platforms}
Table~\ref{tab:mobile_platform_comparison} shows that the benefit of caching is consistent across both smartphones. On iPhone 16 Pro, caching reduces latency from 4.342\,ms to 0.184\,ms; on Xiaomi 14, it reduces latency from 16.237\,ms to 0.547\,ms. In both cases, the cached version remains close to Sensor-Only, while the uncached version is much slower. Memory changes only slightly between cached and uncached variants on the same platform, indicating that caching mainly reduces repeated context-encoding cost rather than runtime footprint.

\begin{table}
    \centering
    \footnotesize
    \setlength{\tabcolsep}{4pt}
    \renewcommand{\arraystretch}{1.1}
    \begin{tabular}{lccccc}
        \toprule
        \textbf{Method} &
        \textbf{\makecell{Params\\(M)}} &
        \textbf{\makecell{Size\\(MB)}} &
        \textbf{\makecell{Latency\\(ms)}} &
        \textbf{\makecell{Memory\\(MB)}} &
        \textbf{\makecell{Energy\\(mJ)}} \\
        \midrule
        Sensor-Only & 0.102 & 0.204 & 0.378 & 129.5 & 10390.9 \\
        Fix-Concat & 0.177 & 0.349 & 0.358 & 133.7 & 2836.1 \\
        CACTUS & 0.185 & 0.374 & 0.648 & 854.27 & 5515.7 \\
        \framework{} w/o Cache & 0.300 & 0.578 & 16.237 & 474.5 & 4201.6 \\
        \textbf{\framework{} w/ Cache} & \textbf{0.300} & \textbf{0.578} & \textbf{0.547} & \textbf{470.0} & \textbf{3287.3} \\
        \bottomrule
    \end{tabular}
    \caption{Overhead on Xiaomi 14 for IMU inference.}
    \label{tab:model_comparison}
\end{table}

\begin{table}
    \centering
    \footnotesize
    \setlength{\tabcolsep}{4pt}
    \renewcommand{\arraystretch}{1.1}
    \begin{tabular}{llcc}
        \toprule
        \textbf{Platform} &
        \textbf{Approach} &
        \textbf{\makecell{Latency\\(ms)}} &
        \textbf{\makecell{Memory\\(MB)}} \\
        \midrule
        \multirow{3}{*}{iPhone 16 Pro}
        & Sensor-Only & 0.141 & 19.5 \\
        & \framework{} w/o Cache & 4.342 & 134.5 \\
        & \textbf{\framework{} w/ Cache} & \textbf{0.184} & \textbf{129.9} \\
        \midrule
        \multirow{3}{*}{Xiaomi 14}
        & Sensor-Only & 0.378 & 129.5 \\
        & \framework{} w/o Cache & 16.237 & 474.5 \\
        & \textbf{\framework{} w/ Cache} & \textbf{0.547} & \textbf{470.0} \\
        \bottomrule
    \end{tabular}
    \caption{Inference overhead across mobile platforms.}
    \label{tab:mobile_platform_comparison}
    \vspace{-1em}
\end{table}

\subsection{Real-world Case Study}
\label{sec:case_study}

\textbf{IMU-based HAR with Smartphones under Continuous Context Shifts\footnote{All the data collection was approved by the Institutional Review Board (IRB) of the authors' institution}.} We deploy \framework{} as an iOS application via Core ML~\cite{apple_coreml_docs} on an iPhone 16 Pro. The app captures raw IMU streams at 50\,Hz and performs on-device inference every 0.5\,s using 128-sample windows. We evaluate a continuous sequence with two practical deployment locations, \textit{Left pocket} and \textit{Wrist}, while the user performs \textit{Sitting}, \textit{Standing}, \textit{Jogging}, and \textit{Standing}. We compare \framework{} against a Sensor-Only baseline under the same application pipeline.

\begin{figure}
    \centering
    \setlength{\abovecaptionskip}{0cm}
    \setlength{\belowcaptionskip}{0cm}
    \includegraphics[width=\columnwidth]{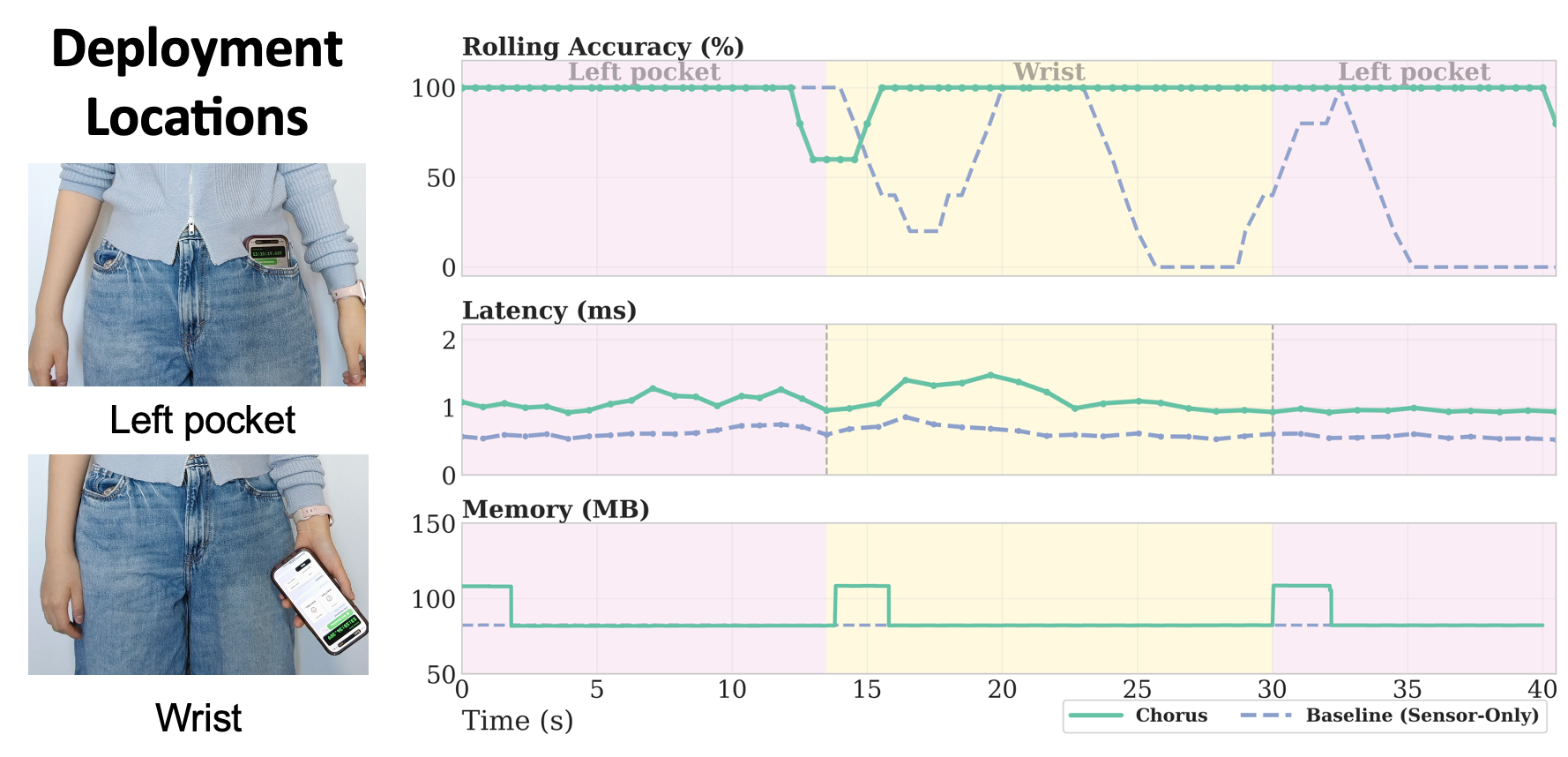}
    \caption{Real-world IMU deployment on iPhone. Left: two practical deployment locations, \textit{Left pocket} and \textit{Wrist}. Right: rolling accuracy, latency, and memory during a continuous context/activity-shift demo.}
    \label{fig:overtime_demo}
\end{figure}

Figure~\ref{fig:overtime_demo} shows that \framework{} remains substantially more stable than the Sensor-Only baseline throughout the sequence. \framework{} achieves 96.3\% mean rolling accuracy, compared with 58.5\% for the baseline, with much lower fluctuation. Although it shows a brief dip around the \textit{Left pocket} $\rightarrow$ \textit{Wrist} transition, it quickly recovers and maintains stable predictions. This gain comes with only a small latency overhead: \framework{} reaches 1.069\,ms mean latency and 1.519\,ms P95, compared with 0.614\,ms and 0.827\,ms for Sensor-Only. Memory spikes are brief and align with context transitions, indicating that dynamic context caching concentrates additional cost at transition points rather than during steady-state inference.

\section{Discussion}

\noindent\textbf{\textcolor{blue}{Generalization beyond observed source-context coverage.}}
\textcolor{blue}{Although \framework{} improves robustness under unseen context shifts, its context representation space is still learned from a finite set of source contexts. 
This calls for stronger effective context representation learning that can capture continuous or compositional context factors, while still aligning these context representations with sensor-space variations.}

\noindent\textbf{\textcolor{blue}{Richer context representations.}}
Our current design focuses on compact, transferable, and sensing-aligned representations derived from short context descriptions, such as placement or environment tags. 
Future work can extend this representation space to richer deployment-time context descriptions, including continuous context attributes, multi-factor context combinations, and uncertainty in context acquisition.

\noindent\textbf{\textcolor{blue}{Efficient integration under dynamic deployment.}}
\framework{} currently integrates context priors through a lightweight gating module and reduces repeated context encoding through dynamic context caching. 
A natural extension is to make efficient context integration more robust under rapidly changing or partially detected contexts, such as noisy context detectors or overlapping context shifts.

\section{Conclusion}
We presented \framework{}, a context-bridged, data-free post-deployment model customization approach for IoT sensing under unseen deployment contexts.
\framework{} learns compact, transferable, and sensing-aligned context representations from unlabeled sensor--context pairs, and integrates them as structured priors through a lightweight gating module and dynamic context caching.
Experiments on IMU sensing, speech enhancement, and WiFi sensing show that \framework{} improves robustness under unseen context shifts by bridging context generalization with sensing-data generalization, while maintaining low inference latency on smartphones and edge devices.


\bibliographystyle{ACM-Reference-Format}
\bibliography{reference}


\clearpage
\appendix
\onecolumn

\end{document}